\documentclass{article}
\usepackage[
  top=2.5cm,
  bottom=2.5cm,
  left=3cm,
  right=3cm
]{geometry}
\usepackage{booktabs}
\usepackage{multirow}
\usepackage{comment}
\usepackage{latexsym}
\usepackage{amssymb}
\usepackage{amsmath}
\usepackage{authblk}
\usepackage{mathtools}
\usepackage{subcaption}
\usepackage{amsthm}
\usepackage{booktabs}
\usepackage{enumitem}
\usepackage{graphicx}
\usepackage{color}
\usepackage{subcaption}
\usepackage{hyperref}
\usepackage{arydshln}
\newtheorem{theorem}{Theorem}
\newtheorem{lemma}{Lemma}
\newtheorem{corollary}{Corollary}

\newtheorem{definition}{Definition}
\newtheorem{assumption}{Assumption}
\usepackage{algorithm}
\usepackage{algorithmicx}
\usepackage{algpseudocode}

\date{}

\begin{document}

\title{Convergence Analysis of Aggregation-Broadcast in LoRA-enabled Distributed Fine-Tuning}


\author[1]{Xin Chen}
\author[2]{Shuaijun Chen}
\author[3]{Omid Tavallaie}
\author[2]{Nguyen Tran}
\author[1]{Shuhuang Xiang}
\author[2]{Albert Zomaya}

\affil[1]{School of Mathematics and Statistics, Central South University, China\\
\texttt{\{chenxin, xiangsh\}@csu.edu.cn
}}

\affil[2]{School of Computer Science, The University of Sydney, Australia\\
\texttt{\{shuaijun.chen, nguyen.tran, albert.zomaya\}@sydney.edu.au
}}

\affil[3]{Department of Engineering Science, University of Oxford, England\\
\texttt{\{omid.tavallaie\}@eng.ox.ac.uk
}}


\maketitle

\begin{abstract}
Federated Learning (FL) enables collaborative model training across decentralized data sources while preserving data privacy. However, the growing size of Machine Learning (ML) models poses communication and computation challenges in FL. Low-Rank Adaptation (LoRA) has recently been introduced into FL as an efficient fine-tuning method, reducing communication overhead by updating only a small number of trainable parameters. Despite its effectiveness, how to aggregate LoRA-updated local models on the server remains a critical and understudied problem. In this paper, we provide a unified convergence analysis for LoRA-based FL. We first categories the current aggregation method into two major type: \textbf{S}um-\textbf{P}roduct (SP) and \textbf{P}roduct-\textbf{S}um (PS). Then we formally define the \textbf{Aggregation-Broadcast Operator (ABO)} and derive both weak and strong convergence condition under mild assumptions. Furthermore, we present both weak and strong convergence condition that guarantee convergence of the local model and the global model respectively. These theoretical analyze offer a principled understanding of various aggregation strategies. Notably, we prove that the SP and PS aggregation methods satisfy the weak and strong convergence condition respectively, but differ in their ability to achieve the optimal convergence rate. Extensive experiments on standard benchmarks validate our theoretical findings.


\end{abstract}



\section{Introduction}
Federated Learning (FL) has emerged as a promising framework for training machine learning models across decentralized data sources while preserving data privacy\cite{mcmahan2017communication,kairouz2021advances}. However, the growing complexity of modern deep neural networks poses significant challenges for communication efficiency and resource-constrained devices, which are central concerns in FL \cite{9052677}. Low-Rank Adaptation (LoRA), a parameter-efficient fine-tuning technique initially proposed for large-scale language models \cite{hu2022lora}, has recently attracted increasing interest in FL due to its ability to reduce communication overhead by updating only a small subset of trainable parameters. The core idea of LoRA is to constrain the weight update on the model by a low rank decomposition:

\begin{equation}
    W' = W_0 + \Delta W, \quad
    \Delta W = BA, \quad
\end{equation}
Where $B \in \mathbb{R}^{d \times r}$ and $A \in \mathbb{R}^{r \times n}$, with $r \ll d$. By training only $B$ and $A$, LoRA significantly reduces the number of trainable parameters while maintaining model performance. Using LoRA in a federated learning setting is an effective and resource-efficient strategy. The global objective of LoRA in FL can be express as:
\begin{equation}
    \arg\min_{A,B} \frac{1}{m}\sum_{i=1}^m 
    \mathbb{E}_{(x,y)\sim P_{XY}^{(i)}} 
    \left[
        \mathcal{L}\left(W_0 + AB;\, (x, y)\right)
    \right].
\end{equation}
where \( m \) denotes the number of clients, and \((x,y) \sim P_{XY}^{(i)}\) indicates that the local data of client \(i\) follows the distribution \(P_{XY}^{(i)}\). By leveraging LoRA adapters, clients can fine-tune large foundation models with minimal computational overhead. Because only the low-rank adapter matrices need to be communicated with the central server, this approach greatly reduces communication overhead. Compared to full-parameter fine-tuning, LoRA offers a more scalable and efficient solution for improving model performance in collaborative learning environments.

Despite its advantages, integrating LoRA into federated learning introduces new challenges, particularly in how the locally updated low-rank parameters should be aggregated on the server side\cite{yang2025federated}. Unlike traditional FL, where full model weights or gradients are averaged directly, LoRA-based training requires the design of specialized aggregation strategies that respect the low-rank structure. In recent studies, multiple LoRA aggregation methods have been proposed, which we broadly classify into the following categories in this paper:
\paragraph{Sum-Product-Type(SP) Aggregation Method.} This method is referred to as the SP method throughout the paper. Such a method is also referred to as the ideal aggregation method, as it shares the same form as FedAvg\cite{mcmahan2017communication}. Its aggregation form is as follows:

\begin{align}
    \Delta W = \frac{1}{m}\sum_{i=1}^mB_iA_i
    \label{PS_aggregation}
\end{align}

This aggregation form can unify several recent methods. For example, FlexLoRA\cite{bai2024federated} was the first to aggregate local modal to server by Eq.~(\ref{PS_aggregation}), and then broadcast by Singular Value Decomposition(SVD). FedIT~\cite{10447454} uploads locally fine-tuned LoRA parameters from each client, which are then aggregated on the server using FedAvg to update the global model. FLoRA~\cite{wang2024flora}, a stacking-based LoRA aggregation method, further improves this process by reducing the impact of noise during aggregation.

\paragraph{Product-Sum-Type(PS) Aggregation Method.} This method is referred to as the PS aggregation method throughout the paper. It's aggregation form is as follows:
\begin{align}
    \Delta W = (\frac{1}{m}\sum_{i=1}^mB_i)(\frac{1}{m}\sum_{i=1}^mA_i)
    \label{SP_aggregation}
\end{align}
This form encompasses several existing methods, such as Zero-Padding\cite{cho2023heterogeneous} and RBLA \cite{chen2024rbla} for Heterogeneous LoRA aggregation. FFA-LoRA\cite{sun2024improving}, which freezing LoRA matrix $A_i=A_0$ and only and aggregate LoRA Matrix $B_i$. Moreover, RoLoRA\cite{chen2024robust} employed an alternating form, which only learn aggregate $B_i$ in odd round and $A_i$ in even round. Different from these, FedSA-LoRA\cite{guo2025selective} update and learn both $B_i$ and $A_i$, but only $A_i$ Matrix are shared for aggregation for learning general knowledge, and save $B_i$ locally for capturing client-specific knowledge. 

\paragraph{Other Aggregation Method.} Of course, some aggregation methods cannot be easily categorized as either SP or PS type aggregation methods, such as FedInc\cite{qin2024fedinc}, which proposed a clustering-based aggregation method, enabling more fine-grained and adaptive aggregation. FedEx-LoRA\cite{singhal2024fedexlora} and LoRA-fair\cite{bian2024lora} introduce correction terms during aggregation to make the results closer to the SP aggregation(idea aggregation). CoLR\cite{nguyen2024towards} adopts a hybrid aggregation strategy by locally learning matrix A while globally sharing matrix B through server-side decomposition. $\text{LoRA-A}^2$\cite{koo2024towards} employs alternating minimization with adaptive rank selection to reduce communication costs by focusing on the most important LoRA ranks.

\paragraph{Motivation}
    Although the two methods mentioned above are now widely used, their underlying mechanisms remain unclear, as we still face the following questions:
\begin{itemize}
    \item The SP aggregation method is referred to as the ideal aggregation method\cite{yang2025federated,guo2025selective}, but is it truly the fastest in terms of convergence speed?
    \item The PS aggregation method is widely adopted\cite{cho2023heterogeneous,chen2024rbla,chen2024robust}, but can it really guarantee convergence for the global model? What's the difference between SP and PS aggregation method in terms of convergence speed? 
    \item For more general aggregation algorithms, under what conditions can they guarantee the convergence of the global model?
\end{itemize}
Addressing these questions is of both theoretical and practical significance. From a theoretical standpoint, a unified convergence analysis helps us understand the fundamental principles that govern the success or failure of various LoRA aggregation strategies. It also provides a rigorous framework to compare different methods on equal footing. From a practical perspective, identifying the conditions that guarantee convergence can guide the design of more effective aggregation algorithms, enabling faster training and better performance in real-world federated learning scenarios.

\paragraph{Contribution}
Our research is primarily motivated by addressing the above questions, and on this basis, we have proposed some more general conclusions. We summarize our contributions as follows:
\begin{itemize}
    \item We formally define the aggregation-broadcast operator. Under certain assumptions, we establish both weak and strong convergence condition. We prove that when weak convergence conditions are satisfied, the Aggregated Broadcasting Operator (ABO) ensures convergence of local models in the LoRA subspace at a rate of $O(1/\sqrt{T})$, when strong convergence conditions are met, it guarantees convergence of global models in the same subspace at the same rate.

    \item Especially, We prove that the SP aggregation-broadcast operator satisfies the weak convergence condition but cannot achieve the optimal convergence rate due to broadcast errors, whereas the PS operator satisfies the strong convergence condition and achieves both global convergence and the optimal convergence rate.

    \item We perform comprehensive empirical studies to validate our theoretical findings. In particular, we investigate the effects of LoRA rank and the number of local training epochs on the convergence behavior of PS and SP aggregation methods, demonstrating strong alignment with our analytical predictions.

\end{itemize}

\paragraph{Paper organization} In Section~\ref{related_work}, we introduce the related work about FL, LoRA and LoRA-based FL. In Section~\ref{aggregation-broadcast-operator}, we introduced the core concept of this work—the \textit{Aggregation-Broadcast Operator} (ABO), which provides a unified view of LoRA parameter aggregation and broadcast in FL. In Section~\ref{section:Analyze}, we established a general convergence condition applicable to any ABO, along with three practical sufficient conditions. Each condition is accompanied by a theoretical convergence guarantee and an associated optimality condition. Finally, in Section~\ref{experiment}, we conducted extensive experiments to validate our theoretical results and further analyzed how different LoRA ranks and local training epochs influence the behavior of PS and SP aggregation methods.

\section{Related Work} \label{related_work}
\paragraph{Federated Learning}
McMahan et al. introduced FedAvg in \cite{mcmahan2017communication} as a decentralized and privacy-aware model training approach. Since then, numerous works have addressed the challenges of non-IID data \cite{zhao2018federated}, communication efficiency \cite{konevcny2016federated}, and personalization \cite{smith2017federated}. Several algorithms have been proposed to improve optimization in heterogeneous settings, including FedProx \cite{li2020federated}, SCAFFOLD \cite{karimireddy2020scaffold}, and MOON \cite{li2021model}. Recent efforts also explore fairness \cite{li2019fair} and adaptive aggregation \cite{wang2020tackling} to balance performance across clients.

\paragraph{LoRA}
Low-Rank Adaptation (LoRA) have emerged as efficient Parameter-Efficient-Fine-Tuning (PEFT) method for Large Language Models (LLMs)\cite{hu2021lora}. Early work such as Universal Language Model Fine-tuning (ULMFiT) also explored efficient adaptation methods to reduce overhead \cite{howard2018universal}. Based on this, QA-LoRA \cite{xu2023qa} introduces quantization-aware adaptation by integrating LoRA with low-bit quantization to further reduce memory usage. Similarly, QLoRA \cite{dettmers2024qLoRA} extends this idea by using 4-bit quantized LoRA adapters, achieving competitive performance with significantly lower memory footprint. In vanilla LoRA, the model rank requires manual configuration, to solve this issue, AdaLoRA \cite{zhang2023adalora} and AutoLoRA\cite{zhang-etal-2024-autolora} automates rank selection process, allowing the model to adaptively allocate parameters where most needed. In theoretical analysis, Sadhika et al.\cite{malladi2023kernel} investigate LoRA fine-tuning through the lens of kernel theory and show that, in the lazy training regime, its behavior closely mirrors that of full fine-tuning.

\paragraph{LoRA-based Federated Learning}

Based on the above advancements, numerous LoRA-based methods have been proposed for FL \cite{wu2024fedlora, cho2024heterogeneous, yi2023fedlora, bai2024federated, chen2024rbla, zhang2024towards}. These approaches leverage low-rank LoRA adapters in place of full rank local models to significantly reduce communication and computational overhead, while maintaining strong adaptability in heterogeneous environments.

\section{Aggregation-Broadcast Operator}\label{aggregation-broadcast-operator}
\paragraph{Notation} Let \( W_i^{(t)} \), \( B_i^{(t)} \), and \( A_i^{(t)} \) denote the local model parameters and the LoRA adapter matrices for client \( i \) at step \( t \), $(1\le i\le m,1\le t\le T)$. Correspondingly, let \( W^{(t)} \), \( B^{(t)} \), and \( A^{(t)} \) represent the global model parameters at step \( t \). The initial model \( W_0 \) refers to the pretrained global model weight.

We define the set of global synchronization steps $\mathcal{I}_E$ as:

\begin{equation}
    \mathcal{I}_E = \{ nE \mid n \in \mathbb{N}^+ \},
\end{equation}

where \( E \) denotes the communication interval. If \( t+1 \in \mathcal{I}_E \), then step \( t+1 \) corresponds to a communication round.

Let \( \mathcal{L}_i(W_i^{(t)};\xi_ {i,t}) \) be the local loss function for client \( i \) at step t with $\mathbb E\left[\mathcal{L}_i(W_i^{(t)};\xi_ {i,t})\right]=\mathcal{L}_i(W_i^{(t)})$, where $\xi_{i,t}$ is sampled from the $i$-th client's local data uniformly at random at the training step $t$. Define the global objective as the weighted sum of local losses:

\begin{equation}
   \mathcal{L}(W^{(t)};\xi) = \frac{1}{m}\sum_{i=1}^{m} \mathcal{L}_i(W^{(t)};\xi_{i,t}). 
\end{equation}

Since there are many methods to aggregate and broadcast LoRA adapters, which involves completely different operation. To describe the aggregation and broadcast process more generally, we define the aggregate-broadcast operator:
\begin{definition}[\textbf{Aggregation-Broadcast Operator, ABO}]\label{definition_ABF}
    The aggregation-broadcast operator of LoRA matrix $B_i^{(t+1)}$ and $A_i^{(t+1)}$ is defined by:
    \begin{align}
        \mathcal{P}(A_{1\le j\le m}^{(t+1)},B_{1\le j\le m}^{(t+1)})\coloneqq \mathcal{P}(A_1^{t+1},\cdots,A_m^{t+1},B_1^{t+1},\cdots,B_m^{t+1})\nonumber\\
        \mathcal{Q}(A_{1\le j\le m}^{(t+1)},B_{1\le j\le m}^{(t+1)})\coloneqq\mathcal{Q}(A_1^{t+1},\cdots,A_m^{t+1},B_1^{t+1},\cdots,B_m^{t+1})\nonumber
    \end{align}
    If \( t + 1 \in \mathcal{I}_E \), then a communication round is triggered, and each client enters the aggregation-broadcast phase. During this phase, the local LoRA matrix \( B_i^{(t+1)} \) and \(A_i^{(t+1)}\)is updated via aggregate-broadcast operators \( \mathcal{P} \) and \(\mathcal{Q}\), such that:
    \begin{align}
        B_i^{(t+1)} \leftarrow \mathcal{P}(A_{1\le j\le m}^{(t+1)},B_{1\le j\le m}^{(t+1)})\nonumber\\
        A_i^{(t+1)} \leftarrow \mathcal{Q}(A_{1\le j\le m}^{(t+1)},B_{1\le j\le m}^{(t+1)})\nonumber
    \end{align}
    for $1\le i\le m$.
\end{definition}

It is worth noting that the ABO $\mathcal{P}(A_{1\le j\le m}^{(t+1)},B_{1\le j\le m}^{(t+1)})$ and \(\mathcal{Q}(A_{1\le j\le m}^{(t+1)},B_{1\le j\le m}^{(t+1)})\) defined in Definition~\ref{definition_ABF} may depend on all LoRA matrices involved in round \( t+1 \). The resulting matrix represents the new local LoRA update received by each client after both aggregation and broadcasting have been completed. 

The following two examples illustrate the scenario when the aggregation-broadcast operators depend jointly on both local $A_i$ and $B_i$ matrices. And the scenario they depend only on $A_i$ or $B_i$, respectively:


\paragraph{SP-Type Aggregation-Broadcast.} As we can see in Algorithm~\ref{alg:fl_SP_abo}, when FL adopts the PS aggregation, the server first aggregates global model based on local LoRA adapters $A_i$ and $B_i$ by using $\Delta W = \frac{1}{m}\sum_{i=1}^mB_iA_i$, where $m$ is the number of participants. After aggregation, the server applies SVD decompose $\frac{1}{m}\sum_{i=1}^mB_iA_i$ to $\tilde{U} \Sigma \tilde{V}^\top$, and broadcasts the result to all clients. In this process, the updated local LoRA matrices after the aggregation-broadcast step can be expressed as:
\begin{align}
    B_i^{(t+1)} &\leftarrow \mathcal{P}(A_{1\le j\le m}^{(t+1)}, B_{1\le j\le m}^{(t+1)}) = \tilde{U}[:, :r] \Sigma[:r, :r] \label{PS-broadcast1}\\
    A_i^{(t+1)} &\leftarrow \mathcal{Q}(A_{1\le j\le m}^{(t+1)}, B_{1\le j\le m}^{(t+1)}) = \tilde{V}^{\top}[:r, :]
    \label{PS-broadcast2}
\end{align}
According to this formulation, both \( \mathcal{P} \) and \( \mathcal{Q} \) are dependent on the entire set of local matrices \( \{A_j^{(t+1)}\} \) and \( \{B_j^{(t+1)}\} \), for \( 1 \le j \le m \).

\begin{algorithm}[H]
\caption{Federated Learning with \textbf{SP Aggregation-Broadcast Operator}}
\begin{algorithmic}[1] \label{alg:fl_SP_abo}
\Require Number of clients $m$, total steps $T$, local epochs $E$, learning rate $\eta$
\State Initialize LoRA matrices $(A_i^{(0)}, B_i^{(0)}) = (A_{\text{initial}}, B_{\text{initial}})$ for $1 \leq i \leq m$
\For{$t = 0, 1, \dots, T - 1$}
    \For{each client $i \in \{1, \dots, m\}$ in parallel}
        \State Sample mini-batch $\xi_{i,t}$
        \State Local Update:
        \[
        \begin{aligned}
        B_i^{(t+1)} &\gets B_i^{(t)} - \eta \nabla_B \mathcal{L}_i(W_i^{(t)}, \xi_{i,t}) \\
        A_i^{(t+1)} &\gets A_i^{(t)} - \eta \nabla_A \mathcal{L}_i(W_i^{(t)}, \xi_{i,t})
        \end{aligned}
        \]
    \EndFor
    \If{$t+1 \in \mathcal{I}_E$}
        \State \textbf{Server} aggregates:
        \[
        W^{(t)} = W_0 + \sum_{i=1}^m B_i^{(t+1)} A_i^{(t+1)}
        \]
        \State \qquad$[\tilde{U}, \Sigma, \tilde{V}^T] = \text{SVD}(\sum_{i=1}^m B_i^{(t+1)} A_i^{(t+1)})$
        \State \qquad Broadcast $(B_i^{(t+1)}, A_i^{(t+1)}) = (\tilde{U}[:, :r], \Sigma[:r, :r], \tilde{V}^T[:r, :])$ for $1 \leq i \leq m$
    \EndIf
\EndFor
\end{algorithmic}
\end{algorithm}



\paragraph{PS-Type Aggregation-Broadcast.}The process can be seen in Algorithm~\ref{alg:fl_PS_abo}. In contrast, the PS method separately averages \( B_i^{(t+1)} \) and \( A_i^{(t+1)} \), broadcasting the mean of each:
\begin{align}
    B_i^{(t+1)} &\leftarrow \mathcal{P}(A_{1\le j\le m}^{(t+1)},B_{1\le j\le m}^{(t+1)})=\frac{1}{m}\sum_{i=1}^mB_i^{(t+1)}\label{SP-broadcast1}\\
    A_i^{(t+1)} &\leftarrow \mathcal{Q}(A_{1\le j\le m}^{(t+1)},B_{1\le j\le m}^{(t+1)})=\frac{1}{m}\sum_{i=1}^mA_i^{(t+1)}\label{SP-broadcast2}
\end{align}
 Therefore, under PS aggregation, the operator $\mathcal{P}$ responsible for constructing the global update depends solely on the aggregated $B_j^{(t+1)}$, while the operator $\mathcal{Q}$ depends only on the aggregated $A_j^{(t+1)}$, where $1 \leq j \leq m$.

\begin{algorithm}[H]
\caption{Federated Learning with \textbf{PS Aggregation-Broadcast Operator}}
\label{alg:fl_PS_abo}
\begin{algorithmic}[1]
\Require Number of clients $m$, total steps $T$, local epochs $E$, learning rate $\eta$
\State Initialize LoRA matrices $(A_i^{(0)}, B_i^{(0)})=(A_{initial},B_{initial})$ for $1\le i\le m$
\For{$t = 0, 1, \dots, T-1$}
    \For{\textbf{each client} $i \in \{1,\dots,m\}$ \textbf{in parallel}}
        \State Sample mini-batch $\xi_{i,t}$
        \State Local Update:
        \[
        \begin{aligned}
        B_i^{(t+1)} &\gets B_i^{(t)} - \eta \nabla_B \mathcal{L}_i(W_i^{(t)}, \xi_{i,t}) \\
        A_i^{(t+1)} &\gets A_i^{(t)} - \eta \nabla_A \mathcal{L}_i(W_i^{(t)}, \xi_{i,t})
        \end{aligned}
        \]
    \EndFor
    \If{$t+1\in \mathcal{I}_E$}
        \State \textbf{Server} aggregates:
        \[
        W^{(t)} = W_0+(\sum_{i=1}^mB_i^{(t+1)})(\sum_{i=1}^mA_i^{(t+1)})
        \]
        \State \qquad Boradcast $(B_i^{(t+1)}, A_i^{(t+1)}) = (\sum_{j=1}^mB_j^{(t+1)},\sum_{j=1}^mA_j^{(t+1)})$ for $1\le i\le m$
    \EndIf
\EndFor
\end{algorithmic}
\end{algorithm}

Moreover, the one-step update for local LoRA matrix at round $t+1$ by such ABO $\mathcal{P}$ and $\mathcal{Q}$ can be described as follows:
\begin{align}
        \begin{pmatrix}
            B_i^{(t)} \\
            A_i^{(t)}
        \end{pmatrix}
        \xrightarrow{\text{local update}}
        \begin{pmatrix}
            B_i^{(t+1)}=B_i^{(t)}-\eta\nabla_B\mathcal{L}_i(W_i^{(t)};\xi_{i,t}) \\
            A_i^{(t+1)}=A_i^{(t)}-\eta\nabla_A\mathcal{L}_i(W_i^{(t)};\xi_{i,t})
        \end{pmatrix}
        \xrightarrow{\text{if } t+1 \in \mathcal{I}_E}
        \begin{pmatrix}
            \mathcal{P}(A_{1\le j\le m}^{(t+1)},B_{1\le j\le m}^{(t+1)}) \\
            \mathcal{Q}(A_{1\le j\le m}^{(t+1)},B_{1\le j\le m}^{(t+1)})
        \end{pmatrix}.
        \label{local_update}
\end{align}
for $1\le i\le m$.


\section{Analysis}\label{section:Analyze}
\subsection{Assumption}
To facilitate the theoretical analysis of LoRA-based aggregation in federated learning, we begin by introducing several standard assumptions that are widely used in the FL literature. These assumptions ensure that the local objective functions and model updates behave in a stable and analyzable manner. In particular, we assume the smoothness of the loss functions, uniform boundedness of their gradients, and uniform boundedness of the LoRA matrices during the model update process which have been widely adopted in many theoretical analyses of FL~\cite{Li2020On,cho2021client,10.1609/aaai.v33i01.33015693,guo2025selective}.
\begin{assumption}
    $\mathcal{L}_1,\mathcal{L}_2,\cdots,\mathcal{L}_m$  are all L-smooth. For all $V$ and $W$,
    \begin{align}
        \|\nabla\mathcal{L}(V) - \nabla\mathcal{L}(W)\|_F\le L\|V-W\|_F \nonumber
    \end{align}
    it is equivalent to
    \begin{align}
        \mathcal{L}_i(V)\le \mathcal{L}_i(W) + \langle V-W,\nabla \mathcal{L}_i(W) \rangle + \frac{L}{2}\|V-W\|_F^2\nonumber
    \end{align}
    \label{Assumption_L_smooth}
\end{assumption}

\begin{assumption}
    The expected squared norm of stochastic gradient is uniformly bounded, i.e., $\mathbb E\left[||\nabla \mathcal{L}_i(W_i^{(t)};\xi_{i,t})||^2\right] = \|\nabla\mathcal{L}_i(W_i^{(t)})\|^2\le G^2$, for all $i=1,2,\cdots,m$ and $t = 0,\cdots,T-1$. Here $T$ denotes the total number of every client's training steps.
    \label{Assumption_gradient-bounded}
\end{assumption}

\begin{assumption}
    Let $W_i^{(t)} = W_0 + B_i^{(t)}A_i^{(t)}$.There exist constants $C_B >0,C_A>0,c_B>0$ and $c_A>0$ such that $\|B_i^{(t)}\|_F\le C_B$, $\|A_i^{(t)}\|_F \le C_A$ for all $i=1,2,\cdots,m$ and $t = 0,1,\cdots,T-1$.
    \label{Assumption_norm_bounded}
\end{assumption}

\subsection{Convergence Results}
In this section, we start to discuss the convergence result for general  Aggregate-Broadcast Operato(ABO). It's worth to be noticed is that: $(1)$ we use the same learning rate $\eta$ during the whole training process and $(2)$ all the device are active. Under these assumptions, we give the following convergence condition and convergence theorem respectively for arbitrary Aggregate-Broadcast Operator(ABO).

\subsubsection{Weak Convergence Condition}
In what follows, we begin by introducing a convergence condition under which the local model is guaranteed to converge. We refer the condition as the \textbf{weak convergence condition}. This condition imposes mild constraints on the behavior of the  Aggregate-Broadcast Operato(ABO), and serves as the foundation for establishing convergence guarantee.


\begin{definition}[\textbf{Weak Convergence Condition}]\label{Weak_Convergence_Condition}
    The  Aggregation-Broadcast Operators(ABO) of LoRA matrix $B_i^{(t+1)}$ and $A_i^{(t+1)}$ is said to satisfy the Weak Convergence Condition if there exist a constant $R>0$ such that:
    \begin{align}
        \mathbb{E}\left[\frac{1}{m}\sum_{i=1}^m\|\mathcal{P}(A_{1\le j\le m}^{(t+1)},B_{1\le j\le m}^{(t+1)})\mathcal{Q}(A_{1\le j\le m}^{(t+1)},B_{1\le j\le m}^{(t+1)})-B_i^{(t+1)}A_i^{(t+1)}\|_F^2\right] &\le R^2\eta^2
        \label{upperbound_R}
    \end{align}
    for $1\le t\le T$, where $\eta$ is learning rate.
\end{definition}

Based on the convergence condition introduced above, we are now in a position to analyze the convergence behavior of the model under this setting. This is a general convergence theorem for arbitrary ABO satisfy the convergence condition.
\begin{theorem}[\textbf{Weak Convergence Theorem}]\label{Weak_Convergence_Theorem}
     Let Assumption \ref{Assumption_L_smooth}, \ref{Assumption_gradient-bounded} and \ref{Assumption_norm_bounded} hold. If the ABO satisfies the Weak Convergence Condition in Definition~\ref{Weak_Convergence_Condition}, the update for local LoRA matrix followed by Eq.~(\ref{local_update}). Then, for a learning rate $\eta>\xi>0$ for some $\xi>0$, the gradient of the local loss in expectation satisfy:
     \begin{align}
         \frac{1}{mT}\sum_{i=1}^m\sum_{t=1}^T(\mathbb{E}\left[\|\nabla_B L_i(W_i^{{(t)}})\|_F^2\right]+\mathbb{E}\left[\|\nabla_A L_i(W_i^{{(t)}})\|_F^2\right])
        \le 2\sqrt{\frac{DM}{T}}
         \label{convergence_global_R}
     \end{align}
     where $T$ is the total number of communication round, $\mathcal{L}_i(W_i^{0})-\mathcal{L}_i(W_i^*)\le D$ for $\forall i$, $\frac{3}{2}L(\eta^2C_A^2C_B^2G^4+C_A^4G^2+C_B^2G^2)\eta^2+C_AC_BG^3\eta^2+\frac{L}{2}R^2\eta^2+\frac{1}{2}(R^2+G^2)\eta\le M\eta^2$.
     \label{thm_aggregation}
\end{theorem}

The detailed proof of Theorem~\ref{Weak_Convergence_Theorem} is provided in the Appendix.~\ref{Appendix_pf_thm1}. Theorem~\ref{Weak_Convergence_Theorem} establishes that if the Aggregation-Broadcast Operator (ABO) satisfies the weak convergence condition, the local model converges to a stationary point within the subspace spanned by $B$ and $A$, achieving a rate of $\mathcal{O}(1/\sqrt{T})$. Moreover, the convergence can be accelerated by reducing the values of $M$ and $N$, both of which increase as $R$ increases. As a result, the upper bound $R$ in Definition~\ref{Weak_Convergence_Condition}, Eq.~(\ref{upperbound_R}) directly affects the convergence speed: a larger $R$ leads to slower convergence, whereas a smaller $R$ results in faster convergence. 

It is known from the convergence condition in Definition~\ref{Weak_Convergence_Condition} that minimize $R$ is equivalent to solving the following optimal problem:
\begin{align}
    \min_{\mathcal{P,Q}} \max_{1\le t\le T}\mathbb{E}\left[ \frac{1}{m}\sum_{i=1}^m\|\mathcal{P}(A_{1\le j\le m}^{(t+1)},B_{1\le j\le m}^{(t+1)})\mathcal{Q}(A_{1\le j\le m}^{(t+1)},B_{1\le j\le m}^{(t+1)})-B_i^{(t+1)}A_i^{(t+1)}\|_F^2\right]
\end{align}
In particular, if we ignore the dependency on the round $t$ in Eq.~(\ref{upperbound_R}), the objective of minimizing $R$ further reduces to the following simplified problem:
\begin{align}
    \min_{\mathcal{P,Q}}\mathbb{E}\left[\frac{1}{m}\sum_{i=1}^m\|\mathcal{P}(A_{1\le j\le m}^{(t+1)},B_{1\le j\le m}^{(t+1)})\mathcal{Q}(A_{1\le j\le m}^{(t+1)},B_{1\le j\le m}^{(t+1)})-B_i^{(t+1)}A_i^{(t+1)}\|_F^2\right]
    \label{optimization_problem_convergenc_condition}
\end{align}
By solving the convex-optimal problem~\ref{optimization_problem_convergenc_condition}, we can get:
\begin{corollary}\label{corollary_weak_optimal}
    The Aggregation-Broadcast Operators(ABO) $\mathcal{P}$ and $\mathcal{Q}$, can satisfy the convergence condition and achieve the optimal convergence rate of the global model showed in Theorem~\ref{Weak_Convergence_Theorem} if the following equation holds:
    \begin{align}
        \mathcal{P}(A_{1\le j\le m}^{(t+1)},B_{1\le j\le m}^{(t+1)})\mathcal{Q}(A_{1\le j\le m}^{(t+1)},B_{1\le j\le m}^{(t+1)}) = \frac{1}{m}\sum_{i=1}^m B_i^{(t+1)}A_i^{(t+1)}
        \label{weak_optimal_PQ}
    \end{align}
    for $1\le t\le T$, where $R^2 = 8E^2G^2(C_A^4+C_B^4)$.
\end{corollary}
The proof of this corollary can be seen in Appendix~\ref{Appendix_pf_corollary1}. We refer the Eq.~(\ref{weak_optimal_PQ}) as the \textbf{optimality condition} under the weak convergence condition. It's obvious that the SP Aggregation Method (As shown in Eq.~(\ref{PS_aggregation})) satisfy the Eq.~(\ref{weak_optimal_PQ}) in Corollary.~\ref{corollary_weak_optimal} during aggregation phase. However, issues arise during the broadcast phase. As showed in Eq.~(\ref{PS-broadcast1}) and Eq.~(\ref{PS-broadcast2}):
\begin{align}
    \mathcal{P}(A_{1\le j\le m}^{(t+1)}, B_{1\le j\le m}^{(t+1)}) &= \tilde{U}[:, :r] \Sigma[:r, :r] \nonumber\\
    \mathcal{Q}(A_{1\le j\le m}^{(t+1)}, B_{1\le j\le m}^{(t+1)}) &= \tilde{V}^\top[:r, :] \nonumber
\end{align}
It means that
\begin{align}
    \mathcal{P}(A_{1\le j\le m}^{(t+1)}, B_{1\le j\le m}^{(t+1)})\mathcal{Q}(A_{1\le j\le m}^{(t+1)}, B_{1\le j\le m}^{(t+1)}) = \tilde{U}[:, :r] \Sigma[:r, :r]\tilde{V}^{\top}[:r, :] \ne \frac{1}{m}\sum_{i=1}^mB_i^{(t+1)}A_i^{(t+1)}
    \label{boradcast_err}
\end{align}
These results indicate that the SP aggregate-broadcast strategy cannot achieve the optimal convergence rate due to the broadcast error, also referred to as broadcast loss. Notably, this error becomes more significant as the LoRA rank $r$ decreases, thereby further slowing down the convergence of the global model, as seen in fig.~\ref{fig:mnist_sp_homo_rank}. The convergence reaches its optimal rate only when the LoRA rank $r$ equals the rank of the full model, in which case the adaptation process becomes equivalent to full-model training. This, however, deviates from the core motivation of parameter-efficient fine-tuning, that is, updating only a small subset of parameters instead of the entire model. A more comprehensive analysis of this phenomenon will be presented in Section~\ref{section_diff_rank}.


\subsubsection{Sufficient Condition for Weak Convergence}
While the weak convergence condition provides a general theoretical guarantee for local model convergence, it may not be straightforward to verify or enforce in practice. To address this, we next present a practical and verifiable sufficient condition that ensure the weak convergence condition is satisfied. 

\begin{theorem}[\textbf{Sufficient Condition 1}]\label{sufficient_Condition1}
    The Aggregation-Broadcast Operators(ABO) of LoRA matrix $B_i^{(t+1)}$ and $A_i^{(t+1)}$ satisfy the Convergence Condition if there exist a constant $R>0$ such that:
    \begin{align}
        \mathbb{E}\left[\|\mathcal{P}(A_{1\le j\le m}^{(t+1)},B_{1\le j\le m}^{(t+1)})\mathcal{Q}(A_{1\le j\le m}^{(t+1)},B_{1\le j\le m}^{(t+1)})-B_i^{(t+1)}A_i^{(t+1)}\|_F^2 \right]&\le R^2\eta^2
    \end{align}
    for $1\le i\le m$, $1\le t\le T$. Where $\eta$ is learning rate.
\end{theorem}
Compared to the Weak Convergence Condition in Definition~\ref{Weak_Convergence_Condition},  Theorem~\ref{sufficient_Condition1} provides a more practical and easily verifiable sufficient condition. This theorem can be directly proven by:
\begin{align}
    \frac{1}{m}\sum_{i=1}^m\|\mathcal{P}(A_{1\le j\le m}^{(t+1)},B_{1\le j\le m}^{(t+1)})\mathcal{Q}(A_{1\le j\le m}^{(t+1)},B_{1\le j\le m}^{(t+1)})-B_i^{(t+1)}A_i^{(t+1)}\|_F^2 &\le \frac{1}{m}\sum_{i=1}^mR^2\eta^2\le R^2\eta^2
\end{align}
Based on Theorem~\ref{sufficient_Condition1}, we can immediately derive that the local model also converges at the rate of $\mathcal{O}(1/\sqrt{T})$, as showed in Eq.~(\ref{optimization_problem_convergenc_condition}). Moreover, minimizing the constant $R$ in Theorem~\ref{sufficient_Condition1} leads to improved convergence. In fact, this minimization is equivalent to solving the following optimization problem:
\begin{align}
    \min_{\mathcal{P},\mathcal{Q}} \max_{1\le i\le m,1\le t\le T}\mathbb{E}\left[ \|\mathcal{P}(A_{1\le j\le m}^{(t+1)},B_{1\le j\le m}^{(t+1)})\mathcal{Q}(A_{1\le j\le m}^{(t+1)},B_{1\le j\le m}^{(t+1)})-B_i^{(t+1)}A_i^{(t+1)}\|_F^2\right]
\end{align}
In particular, if we ignore the dependency on the communication round $t$, the problem simplifies to the following form:
\begin{align}
     \min_{\mathcal{P},\mathcal{Q}} \max_{1\le i\le m}\mathbb{E}\left[ \|\mathcal{P}(A_{1\le j\le m}^{(t+1)},B_{1\le j\le m}^{(t+1)})\mathcal{Q}(A_{1\le j\le m}^{(t+1)},B_{1\le j\le m}^{(t+1)})-B_i^{(t+1)}A_i^{(t+1)}\|_F^2\right]
\end{align}
This formulation provides a clear objective for designing effective aggregation-broadcast operators by minimizing the bound $R$.

\subsubsection{Strong Convergence Condition}
While the weak convergence condition ensures the convergence of local models, it does not necessarily guarantee the convergence of the global model. In practice, the ultimate objective of federated learning is to achieve stable and efficient convergence of the global model. Therefore, this, we introduce a stronger convergence condition together with its corresponding theorem. This strong convergence condition imposes stricter requirements on the aggregation-broadcast operator but provides the stronger guarantee that the global model will converge in the subspace spanned by $B$ and $A$.

\begin{definition}[\textbf{Strong Convergence Condition}]\label{sufficient_Condition2}
    The Aggregation-Broadcast Operators(ABO) of LoRA matrix $B_i^{(t+1)}$ and $A_i^{(t+1)}$ is said satisfy the Strong Convergence Condition if there exist some constant $P>0$ and $Q>0$ such that:
    \begin{align}
        \mathbb{E}\left[\frac{1}{m}\sum_{i=1}^m\|\mathcal{P}(A_{1\le j\le m}^{(t+1)},B_{1\le j\le m}^{(t+1)})-B_i^{(t+1)}\|_F^2\right]\le P^2\eta^2 \label{sum_PB}\\
        \mathbb{E}\left[\frac{1}{m}\sum_{i=1}^m\|\mathcal{Q}(A_{1\le j\le m}^{(t+1)},B_{1\le j\le m}^{(t+1)})-A_i^{(t+1)}\|_F^2\right]\le Q^2\eta^2 \label{sum_QA}
    \end{align}
    for $1\le i\le m$, $1\le t\le T$.
\end{definition}

We define the global weight in step $t$ as $W^{(t)} = W_0+\mathcal{P}(A_{1\le j\le m}^{(t)},B_{1\le j\le m}^{(t)})\mathcal{Q}(A_{1\le j\le m}^{(t)},B_{1\le j\le m}^{(t)})$. Then we can get the following Strong Convergence Theorem under Strong Convergence Condition.

\begin{theorem}[\textbf{Strong Convergence Theorem}]\label{Strong_Convergence_thm}
    Let Assumption \ref{Assumption_L_smooth}, \ref{Assumption_gradient-bounded} and \ref{Assumption_norm_bounded} hold. If the ABO satisfies the Strong Convergence Condition in Definition~\ref{sufficient_Condition2}, the update for local LoRA matrix followed by Eq.~(\ref{local_update}). Then, for a learning rate $\eta>\xi>0$ for some $\xi>0$, the gradient of the global loss in expectation satisfy:
    \begin{align}
        \qquad\frac{1}{T}\sum_{t=1}^T(\mathbb{E}\left[\|\nabla_B\mathcal{L}(W^{(t)})\|_F^2\right]+\mathbb{E}\left[\|\nabla_A\mathcal{L}(W^{(t)})\|_F^2\right])\le 4\sqrt{\frac{D(M+N)}{T}}
    \end{align}
    for $1\le i\le m$, $1\le t\le T$, where $\mathcal{L}_i(W_0) - \mathcal{L}_i(W_i^*)\le D$, $R^2 = 4P^2Q^2\eta^2+3C_B^2Q^2 +3C_A^2P^2$, $4G^2(Q^2+P^2)\eta^2 +4(C_A^2+C_B^2)L^2R^2\eta^2\le 2N\eta$ and $\frac{3}{2}L(\eta^2C_A^2C_B^2G^4+C_A^4G^2+C_B^2G^2)\eta^2+C_AC_BG^3\eta^2+\frac{L}{2}R^2\eta^2+\frac{1}{2}(R^2+G^2)\eta\le M\eta^2$.
\end{theorem}
The proof of this theorem can be seen in Appendix~\ref{Appendix_pf_sc2}. Similarly, to further simplify the analysis, if we ignore the dependency on the round $t$, minimize $R$ under Strong Convergence Condition in Definition~\ref{sufficient_Condition2} is equivalent to solve the following convex-optimal problem\cite{boyd2004convex}:
\begin{align}
    \min_{\mathcal{P}}\mathbb{E}\left[\frac{1}{m}\sum_{i=1}^m\|\mathcal{P}(A_{1\le j\le m}^{(t+1)},B_{1\le j\le m}^{(t+1)})-B_i^{(t+1)}\|_F^2\right]\label{opt_PB}\\
    \min_{\mathcal{Q}}\mathbb{E}\left[\frac{1}{m}\sum_{i=1}^m\|\mathcal{Q}(A_{1\le j\le m}^{(t+1)},B_{1\le j\le m}^{(t+1)})-A_i^{(t+1)}\|_F^2\right]\label{opt_QA}
\end{align}
By solving the optimal problem we can get the corollary:

\begin{corollary}\label{col_Popt_Qopt}
     The Aggregation-Broadcast Operators(ABO) $\mathcal{P}$ and $\mathcal{Q}$ can satisfy the Strong Convergence Condition in Definition~\ref{sufficient_Condition2} and achieve the optimal convergence rate of the global model showed in Theorem~\ref{Strong_Convergence_thm} if the following equation holds:
     \begin{align}
         \mathcal{P}(A_{1\le j\le m}^{(t+1)},B_{1\le j\le m}^{(t+1)}) &= \frac{1}{m}\sum_{i=1}^m B_i^{(t+1)}
         \label{optmal_PB}\\
         \mathcal{Q}(A_{1\le j\le m}^{(t+1)},B_{1\le j\le m}^{(t+1)}) &= \frac{1}{m}\sum_{i=1}^mA_i^{(t+1)}
         \label{optmal_QA}
     \end{align}
    for $1\le t\le T$, where $P^2 = 4E^2G^2C_A^4$, $Q^2=4E^2G^2C_B^2$, which lead to $R^2 = 64E^4G^4C_A^2C_B^2\eta^2+12E^2G^2(Q^2C_B^4+P^2C_A^4)$.
\end{corollary}
The proof of this corollary can be seen in~\ref{Proof_of_col4}. We refer the Eq.~(\ref{optmal_PB}) and~(\ref{optmal_QA}) as the \textbf{optimality condition} under the Strong Convergence Condition. It is worth noting that the PS Aggregation Method satisfy this optimality condition. As a result, the PS Aggregation Method is capable of achieving the optimal convergence rate of the global model if all client share the same LoRA rank,  which suggests that the PS aggregation method is relatively robust to the choice of LoRA rank, as we can see in fig.~\ref{fig:mnist_ps_homo_rank}. We will provide a more detailed analysis of this issue in Section~\ref{section_diff_rank}.

\subsubsection{Sufficient Condition for Strong Convergence}
We now turn to an another sufficient condition. In contrast to Strong Convergence Condition, it offers a more concise formulation and is more amenable to verification. This makes it particularly useful for practical implementation and analysis.
\begin{theorem}[\textbf{Sufficient Condition 2}]\label{sufficient_Condition3}
    The Aggregation-Broadcast Operators(ABO) of LoRA matrix $B_i^{(t+1)}$ and $A_i^{(t+1)}$ satisfy the  Strong Convergence Condition if there exist some constant $P>0$ and $Q>0$ such that:
    \begin{align}
        \mathbb{E}\left[\|\mathcal{P}(A_{1\le j\le m}^{(t+1)},B_{1\le j\le m}^{(t+1)})-B_i^{(t+1)}\|_F^2\right]\le P^2\eta^2\\
        \mathbb{E}\left[\|\mathcal{Q}(A_{1\le j\le m}^{(t+1)},B_{1\le j\le m}^{(t+1)})-A_i^{(t+1)}\|_F^2\right]\le Q^2\eta^2
    \end{align}
    for $1\le i\le m$, $1\le t\le T$. Where $\eta$ is learning rate.
\end{theorem}
This theorem can be easily proofed by:
\begin{align}
    \mathbb{E}\left[\frac{1}{m}\sum_{i=1}^m\|\mathcal{P}(A_{1\le j\le m}^{(t+1)},B_{1\le j\le m}^{(t+1)})-B_i^{(t+1)}\|_F^2\right]
    \le \frac{1}{m}\sum_{i=1}^m P^2\eta^2 = P^2\eta^2\\
    \mathbb{E}\left[\frac{1}{m}\sum_{i=1}^m\|\mathcal{Q}(A_{1\le j\le m}^{(t+1)},B_{1\le j\le m}^{(t+1)})-A_i^{(t+1)}\|_F^2\right]
    \le \frac{1}{m}\sum_{i=1}^m Q^2\eta^2 = Q^2\eta^2
\end{align}
If we ignore the dependency on the round $t$, minimize $R$ under Sufficient Condition 3 in Theorem~\ref{sufficient_Condition3} is equivalent to solve the following convex-optimal problem:
\begin{align}
    \min_{\mathcal{P}}\max_{1\le i\le m}\sum_{i=1}^m\|\mathcal{P}(A_{1\le j\le m}^{(t+1)},B_{1\le j\le m}^{(t+1)})-B_i^{(t+1)}\|_F^2\label{opt_PB3}\\
    \min_{\mathcal{Q}}\max_{1\le i\le m}\sum_{i=1}^m\|\mathcal{Q}(A_{1\le j\le m}^{(t+1)},B_{1\le j\le m}^{(t+1)})-A_i^{(t+1)}\|_F^2\label{opt_QA3}
\end{align}
\section{Experiment and evaluation}\label{experiment}
\subsection{Experiment Setup}
In our study, we evaluate the effectiveness of the proposed convergence theory using a MLP. The experiments are conducted on the MNIST dataset under a highly non-IID label distribution, where each client holds data samples from a single class. We compare representative methods from the PS and SP paradigms. The detailed experimental settings are summarized below.

\begin{itemize}
\item \textbf{Methods for comparison:} There are many SP and PS methods now. We selected the most classic aggregation method FlexLoRA\cite{bai2024federated} and RBLA\cite{chen2024rbla} to represent SP and PS, respectively.

\item \textbf{Model configurations:}  
The backbone model is a classic Multi-Layer Perceptron (MLP) with three fully-connected layers. To enable parameter-efficient adaptation, we replace each dense layer with a LoRA layer. Specifically, the architecture consists of two hidden LoRA layers with output size 200 and a final output LoRA layer with size 10. The full model architecture is listed in Table.~\ref{table:lora_2nn}, when $\delta=1$, the number of trainable parameters is 199950.

\item \textbf{Non-IID settings:}  
In all our experiments, we simulate a classic label-skewed Non-IID scenario. Each client is assigned all training samples corresponding to a single digit class from the MNIST dataset (e.g., one client only observes samples of digit "3"). This results in a highly heterogeneous data distribution across clients, where no client has access to the full label space. Such a setup closely resembles the pathological Non-IID setting discussed in~\cite{zhao2018federated}, and serves as a challenging benchmark to evaluate the generalization ability of PS and SP scheme under extreme label bias.

\item \textbf{Rank scale ratio $\delta$:}  
To control the adaptation capacity of each LoRA layer, we follow the rank scaling scheme introduced in RBLA~\cite{chen2024rbla}. A global scalar $\delta$ is used to adjust the LoRA rank for each layer proportionally. Concretely, we define the effective rank of each layer as $\max(c \cdot \delta, 1)$, where $c$ is a layer-specific constant. In our model, the constants are 160, 100, and 10 for the first, second, and third layers, respectively. Ensuring that the overall rank budget is smoothly distributed across layers based on their size and functional contribution.

\item \textbf{Experiment rank settings:} In our experiments, the global model's rank ratio $\delta$ is set to $\delta=1$, and rest clients hold the same $\delta$, which can be different under different scenarios.

\end{itemize}

\begin{table}[ht]
\centering
\caption{LoRA-augmented model architecture (rank-scaled).}
\begin{tabular}{|l|c|c|c|}
\hline
\textbf{Layer} & \textbf{Output Shape} & \textbf{Activation} & \textbf{LoRA Rank (A,B)} \\
\hline
Input & (784,) & None & -- \\
\hline
LoRALayer & (200,) & ReLU & $(r_1 = \max(160 \cdot \delta, 1))$ \\
\hline
LoRALayer & (200,) & ReLU & $(r_2 = \max(100 \cdot \delta, 1))$ \\
\hline
LoRALayer & (10,) & Softmax & $(r_3 = \max(10 \cdot \delta, 1))$ \\
\hline
\end{tabular}
\label{table:lora_2nn}
\end{table}

\subsection{Experiment for Different Rank}\label{section_diff_rank}
In the previous section, we theoretically showed that when all clients use the same LoRA rank ratio $r$, the effect of $r$ on SP and PS aggregation differs significantly. Specifically, SP-ABO suffers from broadcast errors that degrade convergence as $r$ decreases, whereas PS aggregation is much less sensitive to this effect. In this section, we empirically validate these theoretical findings and focus on examining how the choice of LoRA rank affects the performance of different aggregation-broadcast strategies. 

In this subsection, we consider two scenarios: a homogeneous setting where all clients share the same LoRA rank \( r \), and a heterogeneous setting where each client is assigned a different rank. In both cases, all clients participate in every communication round. We set the total number of communication rounds to 80, with each client performing 5 local epochs per round. The learning rate $\eta$ is fixed at 0.1 unless otherwise specified.

\begin{figure}[b]
    \centering
    \begin{subfigure}[b]{0.27\textwidth}
        \includegraphics[height=38 mm]{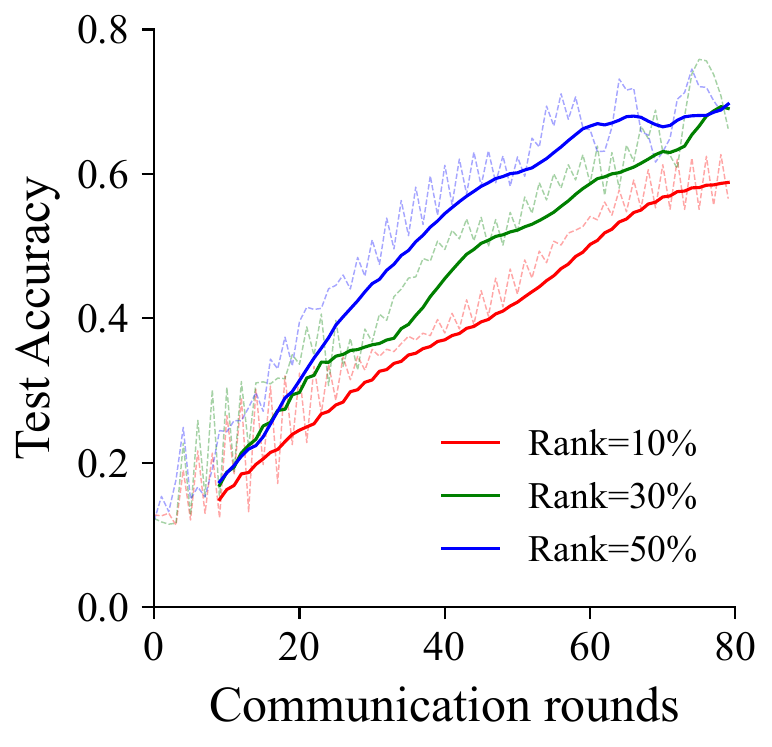}
        \caption{Product-Sum.}
        \label{fig:mnist_ps_homo_rank}
    \end{subfigure}
    \begin{subfigure}[b]{0.27\textwidth}
        \includegraphics[height=38 mm]{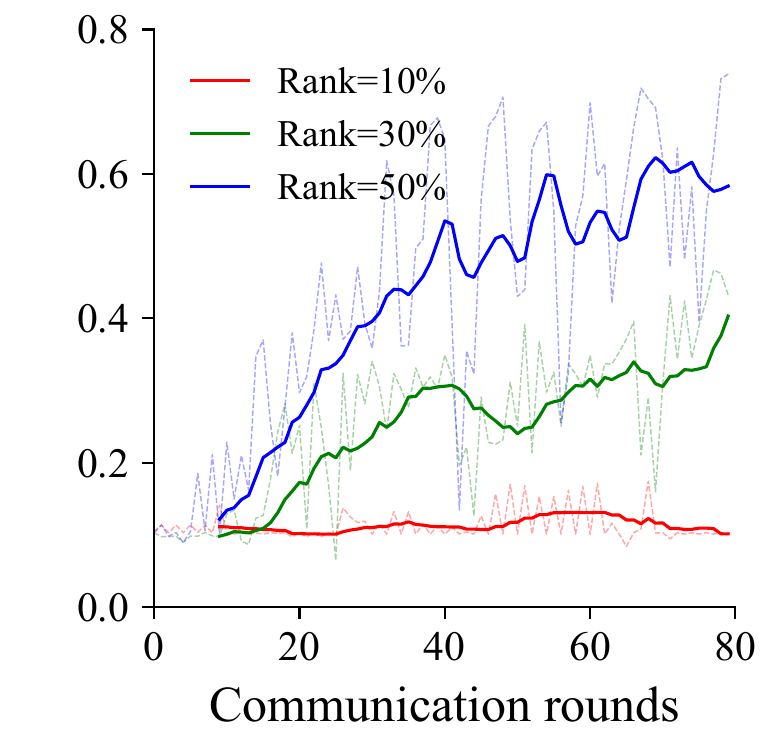}
        \caption{Sum-Product.}
        \label{fig:mnist_sp_homo_rank}
    \end{subfigure}\
    \caption{Comparison of PS and SP of homogeneous ranks on MNIST dataset.}
    \label{fig:ps_sp_homo_rank}
\end{figure}

\paragraph{Homogeneous rank}
Figure~\ref{fig:ps_sp_homo_rank} presents the test accuracy learning curves of PS-ABO and SP-ABO under different homogeneous LoRA ranks, namely $\delta = 0.1$, $0.5$ and $1.0$. Interestingly, this experimental outcome is consistent with our theoretical analysis. As shown in Eq.~(\ref{SP-broadcast1}) and Eq.~(\ref{SP-broadcast2}), the PS-ABO satisfies the optimality condition described in Corollary.~\ref{col_Popt_Qopt}, which guarantees convergence regardless of the LoRA rank. Thus, the results observed in Fig.~\ref{fig:mnist_ps_homo_rank} provide empirical support for the theoretical guarantees established in this paper.This confirms that PS aggregation is relatively robust to the choice of LoRA rank.

In contrast, as shown in Eq.~(\ref{boradcast_err}), due to the presence of broadcast errors introduced by the SVD decomposition, a smaller LoRA rank causes the SP-ABO to deviate further from its optimality condition described in Corollary.~\ref{corollary_weak_optimal}. This theoretical insight is reflected in the empirical results shown in Fig.~\ref{fig:mnist_sp_homo_rank}, which demonstrates that SP-ABO is highly sensitive to the rank value. When the rank is small (e.g., $\delta=0.1$), the model fails to converge due to significant broadcast error. As the rank increases to $\delta=0.3$ and $\delta=0.5$, the convergence improves gradually, supporting our theoretical claim that SP-ABO suffers from larger broadcast-induced errors at lower ranks. These experimental results align well with our theoretical analysis in the previous section.

\subsection{Experiment For Different Epoch}

\begin{figure}[t]
    \centering
    \begin{subfigure}[b]{0.27\textwidth}
        \includegraphics[height=38 mm]{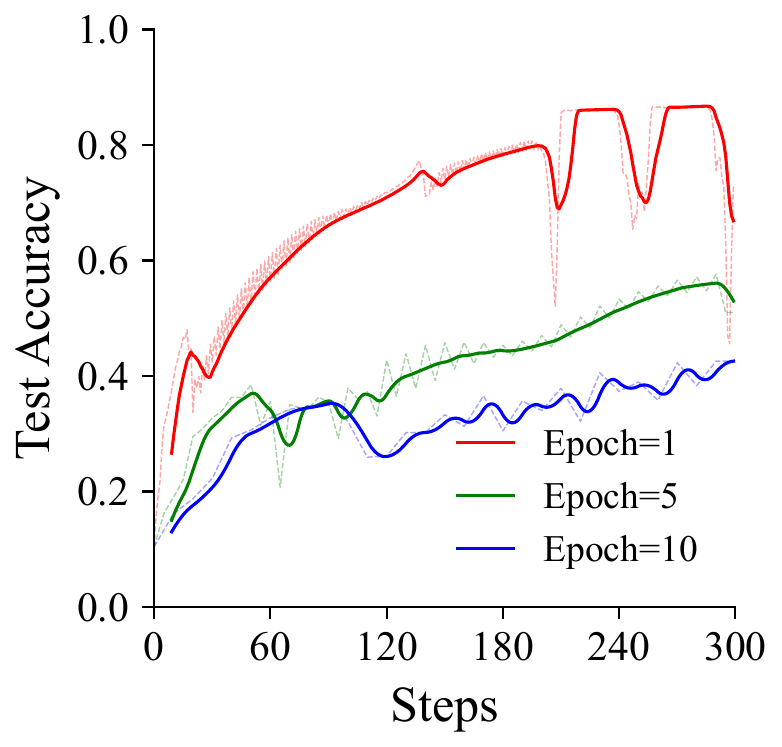}
        \caption{$\delta = 1$.}
        \label{fig:mnist_sp_ps_heterogeneous_client_rank}
    \end{subfigure}
    \begin{subfigure}[b]{0.27\textwidth}
        \includegraphics[height=38 mm]{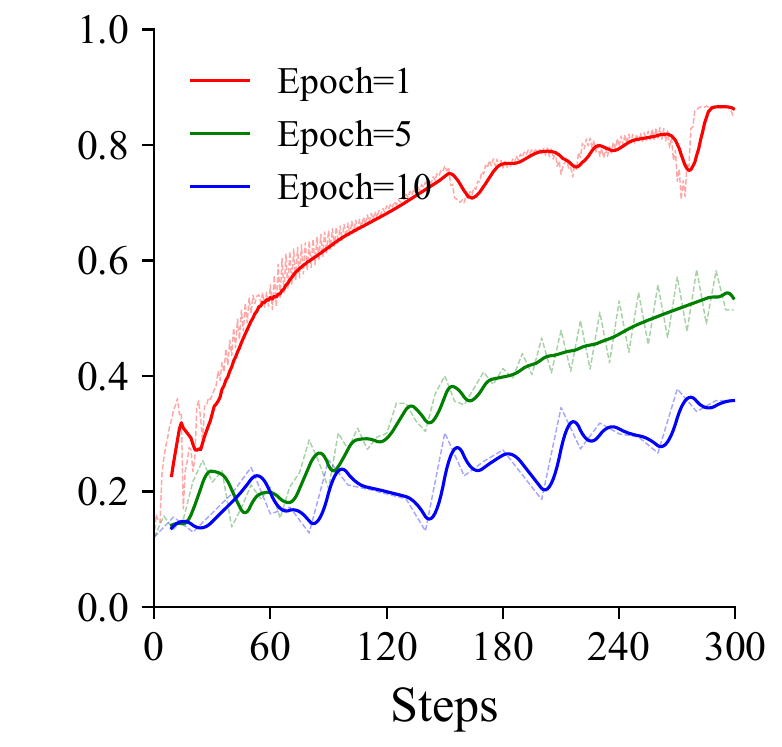}
        \caption{$\delta = 0.5$.}
        \label{fig:fmnist_sp_ps_heterogeneous_client_rank}
    \end{subfigure}
    \begin{subfigure}[b]{0.27\textwidth}
        \includegraphics[height=38 mm]{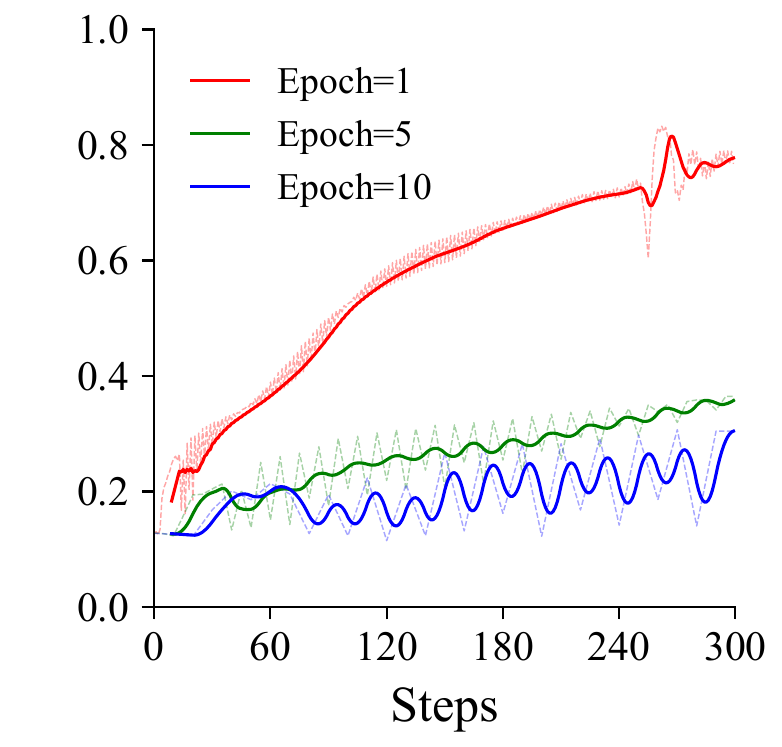}
        \caption{$\delta = 0.1$.}
        \label{fig:fmnist_sp_ps_heterogeneous_client_rank}
    \end{subfigure}\
    \caption{Comparison of PS learning curve on homogeneous ranks.}
    \label{fig:ps_hete_rank_homo_client}
\end{figure}

From Corollary.~\ref{corollary_weak_optimal} and Corollary.~\ref{col_Popt_Qopt} we know that $R^2_{sp} = 8E^2G^2(C_A^4+C_B^4)$, $R^2_{ps} = 64E^4G^4C_A^2C_B^2\eta^2+12E^2G^2(Q^2C_B^4+P^2C_A^4)$. These expressions indicate that, under a fixed total number of total steps $T$ (i.e. total number of epochs), increasing the number of local epochs $E$ per communication round leads to a larger value of $R$, and thus slower convergence for both SP-ABO and PS-ABO. In other words, when $T$ is fixed, performing more local updates between communication rounds degrades overall convergence performance.

\begin{figure}[b]
    \centering
    \begin{subfigure}[b]{0.27\textwidth}
        \includegraphics[height=38 mm]{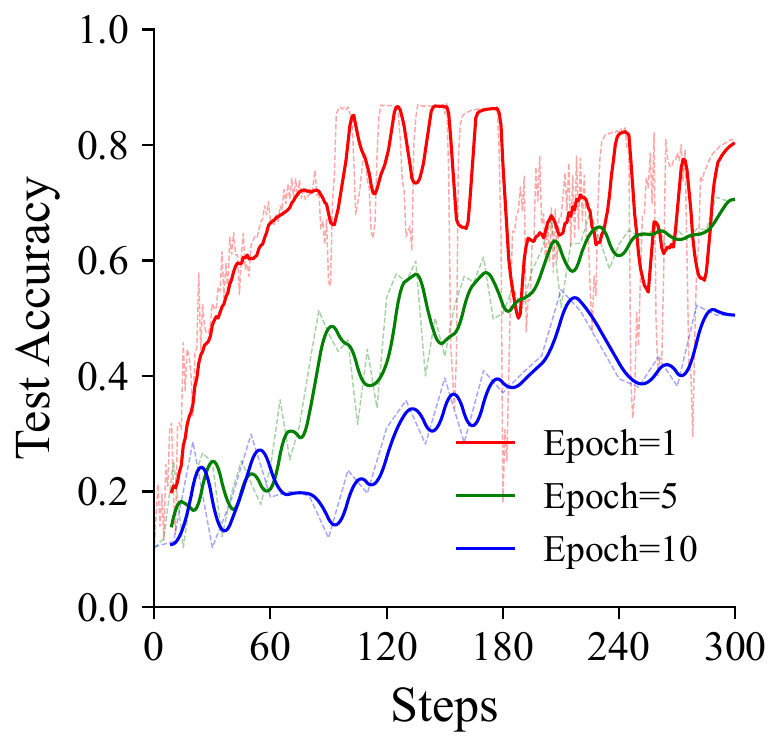}
        \caption{$\delta = 1$.}
        \label{fig:mnist_sp_heterogeneous_client_rank}
    \end{subfigure}
    \begin{subfigure}[b]{0.27\textwidth}
        \includegraphics[height=38 mm]{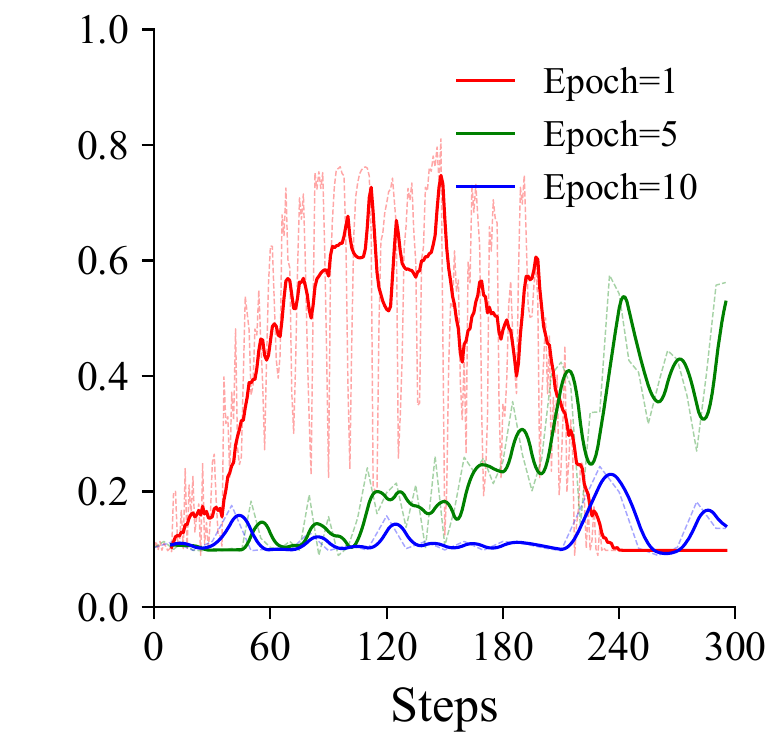}
        \caption{$\delta = 0.5$.}
        \label{fig:mnist_sp_heterogeneous_client_rank}
    \end{subfigure}
    \begin{subfigure}[b]{0.27\textwidth}
        \includegraphics[height=38 mm]{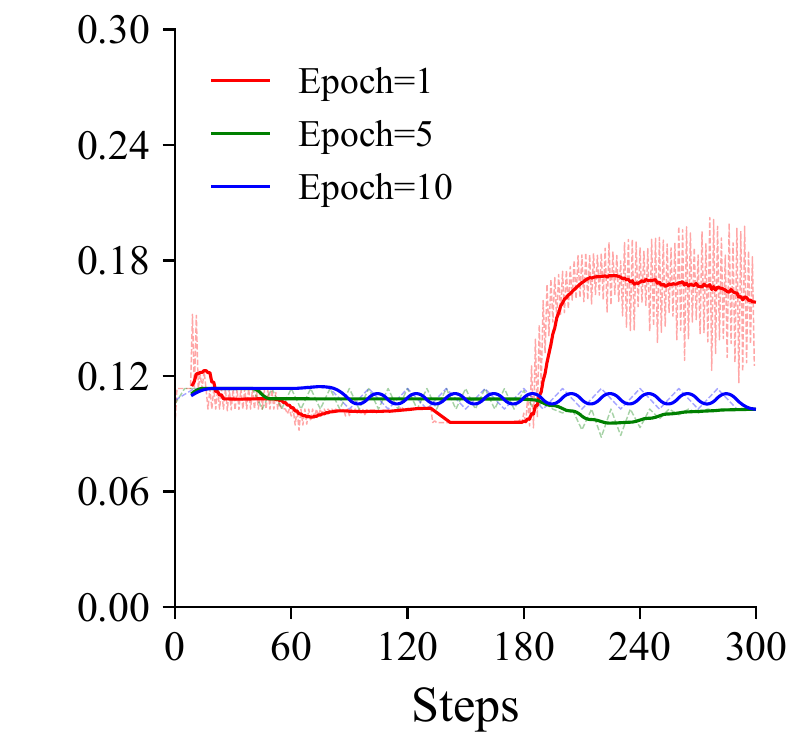}
        \caption{$\delta = 0.1$.}
        \label{fig:mnist_sp_heterogeneous_client_rank}
    \end{subfigure}\
    \caption{Comparison of SP learning curve on homogeneous ranks.}
    \label{fig:sp_hete_rank_homo_client}
\end{figure}

To ensure that constants such as $G$, $C_A$, and $C_B$ remain unchanged, we fix the random seed in all experiments. As a result, we observe that \( R^2_{sp} = \mathcal{O}(E^2) \) and \( R^2_{ps} = \mathcal{O}(E^4) \), which clearly shows that increasing $E$ slows down the convergence of both SP-ABO and PS-ABO, with PS-ABO being more sensitive to the choice of epoch number. In this experiment, we fix the total number of steps to $T = 300$ and set the learning rate \( \eta = 0.05 \). We also assume all clients use the same LoRA rank ratio \( \delta \). We evaluate the performance of SP-ABO and PS-ABO under different values of \( E \) across several homogeneous LoRA ranks. The results are shown in Fig.~\ref{fig:ps_hete_rank_homo_client} and Fig.~\ref{fig:sp_hete_rank_homo_client}, which show the learning curves of the PS and SP-ABO under different LoRA rank ratio \( \delta \in \{1.0, 0.5, 0.1\} \) and local epoch settings \( E \in \{1, 5, 10\} \). 

Table.~\ref{fig:mnist-fmnist-tables} presents the presents the highest test accuracies achieved during training for the PS-ABO and SP-ABO under different LoRA rank ratio \( R \in \{1.0, 0.5, 0.1\} \), and varying numbers of local epochs \( E \in \{1, 5, 10\} \). From the results, we observe two notable patterns: First, the performance of the PS method consistently degrades as the number of local epochs increases, particularly when the rank is fixed. For example, when \( R = 1 \), the accuracy of PS drops from 86.79\% (at \( E = 1 \)) to 42.57\% (at \( E = 10 \)). This confirms our theoretical analysis that PS-ABO is sensitive to the number of local updates due to its stronger dependence on the epoch count \( E \), as reflected in \( R^2_{ps} = \mathcal{O}(E^4) \). Second, the SP method exhibits more robustness to increasing local epochs, but its performance is heavily affected by the choice of LoRA rank. For instance, at \( E = 1 \), the accuracy of SP sharply drops from 87.36\% (when \( \delta = 1 \)) to only 20.22\% (when \( \delta = 0.1 \)). This sensitivity is consistent with our earlier discussion that lower ranks introduce larger broadcast errors in SP-ABO, resulting in greater deviation from its optimality condition.

\begin{table}[t]
    \centering
    \begin{tabular}{|c|c|c|c|c|}
        \hline
        \textbf{Epoch $\backslash$ $\delta$} & \textbf{Method} & \textbf{$\delta = 1$} & \textbf{$\delta = 0.5$} & \textbf{$\delta = 0.1$} \\
        \hline
        \multirow{2}{*}{Epoch = 1} & PS & 86.79\% & 86.75\% & 83.24\% \\
                                   & SP & 87.36\% & 81.05\% & 20.22\% \\
        \hline
        \multirow{2}{*}{Epoch = 5} & PS & 57.62\% & 58.42\% & 36.44\% \\
                                   & SP & 70.95\% & 57.44\% & 11.34\% \\
        \hline
        \multirow{2}{*}{Epoch = 10} & PS & 42.57\% & 37.74\% & 30.7\% \\
                                    & SP & 54.91\% & 24.33\% & 11.47\% \\
        \hline
    \end{tabular}
    \caption{Side-by-side comparison of MNIST and FMNIST tables.}
    \label{fig:mnist-fmnist-tables}
\end{table}

Overall, these results illustrate that PS-ABO is more sensitive to the number of local epochs, while SP-ABO is more sensitive to the LoRA rank. These findings highlight an important trade-off between local computation (epoch count) and parameter efficiency (LoRA rank) in the design of aggregation strategies for federated LoRA training.

\vspace{-4mm}
\section{Conclusion}
\vspace{-2mm}
In this paper, we presented a unified theoretical framework for analyzing the convergence behavior of LoRA-based federated learning. By introducing the concept of aggregation-broadcast operators, we established a general convergence condition along with several sufficient conditions that can be easily verified in practice. Our framework not only provides convergence guarantees for widely used SP-ABO and PS-ABO, but also offers insights into designing new aggregation methods with provable performance. Extensive experiments on standard benchmarks corroborate our theoretical findings. We believe this work lays the foundation for a deeper understanding of LoRA in federated settings and opens up new directions for future research on communication-efficient and convergent model adaptation techniques.

\newpage
\bibliographystyle{plain} 
\bibliography{citation.bib}

\newpage
\appendix
\section{Appendix}
This subsection presents proofs for the theorems and corollaries in the main text, including convergence analysis of local models under arbitrary LoRA aggregation strategies and global model convergence analysis. Specifically, we also provide convergence analysis for the uniform averaging of LoRA matrices.

\subsection{Additional Notation}
Let $W_i^{(t)},B_i^{(t)}$ and $A_i^{(t)}$ denote the local model and local {LoRA matrix} correspondingly for client $i$ at step $t$. Similarly, let $W^{(t)},B^{(t)}$ and $A^{(t)}$ represent the global model parameters at step $t$. Let $W_0$ be the {pre-trained model}. Let $\mathcal{I}_E$ be the set of global synchronization steps, defined as $\mathcal{I}_E = \{nE \mid n\in \mathbb{N^+}\}$ and if $t+1\in \mathcal{I}_E$,  it indicates that the current step is a communication round. For convenience, we denote the aggregation function as follows :
\begin{align}
    P(A_{1\le j\le m}^{(t+1)},B_{1\le j\le m}^{(t+1)})= P(A_1^{(t+1)},\cdots,A_m^{(t+1)},B_1^{(t+1)},\cdots,B_m^{(t+1)})\nonumber\\
    Q(A_{1\le j\le m}^{(t+1)},B_{1\le j\le m}^{(t+1)})=Q(A_1^{(t+1)},\cdots,A_m^{(t+1)},B_1^{(t+1)},\cdots,B_m^{(t+1)})\nonumber
\end{align}
Then the one-step update can be written as:
\begin{align}
    \begin{pmatrix}
        B_i^{(t)} \\
        A_i^{(t)}
    \end{pmatrix}
    \xrightarrow{\text{update of } B_i^{(t)} \text{ and } A_i^{(t)}}
    \begin{pmatrix}
        B_i^{(t+1)} \\
        A_i^{(t+1)}
    \end{pmatrix}
    \xrightarrow{\text{if } t+1 \in \mathcal{I}_E}
    \begin{pmatrix}
        P(A_{1\le j\le m}^{(t+1)},B_{1\le j\le m}^{(t+1)}) \\
        Q(A_{1\le j\le m}^{(t+1)},B_{1\le j\le m}^{(t+1)})
    \end{pmatrix}. \nonumber
\end{align}
We define the following parameters for client $i$ at step $t$ to record information during the local update and model aggregation phases, respectively.:
\begin{align}
    W_i^{(t)} &= W_0 + B_i^{(t)}A_i^{(t)},\nonumber\\
    U_i^{(t)} &= W_0+B_i^{(t+1)}A_i^{(t+1)},\nonumber\\
    V_i^{(t)} &= W_0 + P(A_{1\le j\le m}^{(t+1)},B_{1\le j\le m}^{(t+1)}),Q(A_{1\le j\le m}^{(t+1)}B_{1\le j\le m}^{(t+1)})\nonumber
\end{align}
Here, $U_i^{(t)}$ represents the immediate result of local update, , which captures the information during the local training phase. Similarly, $V_i^{(t)}$ represents the aggregated update,  which captures the information from the aggregation phase. Then we can get:
\begin{align}
    W_i^{(t+1)} =
    \begin{cases}
        U_i^{(t)} & \text{if } t + 1 \notin \mathcal{I}_E, \nonumber\\
        V_i^{(t)} & \text{if } t + 1 \in \mathcal{I}_E.\nonumber
    \end{cases}
\end{align}
\subsection{Key lemmas}
To ensure a concise and clear proof of the theorem, we first provide the following lemmas. Detailed proofs of these lemmas will be presented after completing the proof of Theorem \ref{thm_aggregation}.
\begin{lemma}
    Assume Assumption \ref{Assumption_gradient-bounded} and \ref{Assumption_norm_bounded} hold. We have:
    \begin{align}
        \mathbb{E}\left[\langle U_i^{(t)}-W_i^{(t)}, \nabla_W\mathcal{L}_i(W_i^{(t)}) \rangle_F\right] \le \eta_{i,t}^2C_AC_BG^3 - \eta_{i,t}(\mathbb{E}\left[\|\nabla_B L_i(W_i^{{(t)}})\|_F^2\right]+\mathbb{E}\left[\|\nabla_A L_i(W_i^{{(t)}})\|_F^2\right]) \nonumber
    \end{align}
    \label{lemma_innerproduct}
\end{lemma}
\begin{lemma}
    Assume Assumption \ref{Assumption_gradient-bounded} holds. We have:
    \begin{align}
        \mathbb{E}\left[\|U_i^{(t)} - W_{i}^{(t)}\|_F^2 \right]\le 3\eta_{i,t}^4C_A^2C_B^2G^4 + 3\eta_{i,t}^2C_A^4G^2+3\eta_{i,t}^2C_B^4G^2 \nonumber
    \end{align}
    \label{lemma_dist_UW}
\end{lemma}

\subsection{Proof of theorem \ref{thm_aggregation}} \label{Appendix_pf_thm1}
\begin{proof}
    First, by Assumption~\ref{Assumption_L_smooth}, we have:
    \begin{align}
        \mathcal{L}_i(U_i^{(t)})&\le\mathcal{L}_i(W_i^{(t)})+\langle U_i^{(t)} - W_i^{(t)},\nabla_W\mathcal{L}_i(W_i^{(t)})\rangle_F + \frac{L}{2}\|U_i^{(t)} - W_i^{(t)}\|_F^2 \label{L_smooth_UW}\\
        \mathcal{L}_i(V_i^{(t)})&\le\mathcal{L}_i(U_i^{(t)})+\langle V_i^{(t)} - U_i^{(t)},\nabla_W\mathcal{L}_i(U_i^{(t)}) \rangle_F + \frac{L}{2}\|V_i^{(t)} - U_i^{(t)}\|_F^2 \label{L_smooth_VU}
    \end{align}
    If $t+1\notin \mathcal{I}_E$, by lemma~\ref{lemma_innerproduct} and Eq.~(\ref{L_smooth_UW}) we have:
    \begin{align}
        \mathbb{E}\left[\mathcal{L}_i(W_i^{(t+1)})\right] 
        &= \mathbb{E}\left[\mathcal{L}_i(U_i^{(t)})\right]\nonumber\\
        &\le \mathbb{E}\left[\mathcal{L}_i(W_i^{(t)})\right]+\mathbb{E}\left[\langle U_i^{(t)}-W_i^{(t)},\nabla_W\mathcal{L}_i(W_i^{(t)}) \rangle_F\right]+\frac{L}{2}\mathbb{E}\left[ \|U_i^{(t)} - W_i^{(t)}\|_F^2\right]\nonumber\\
        &\le \mathbb{E}\left[\mathcal{L}_i(W_i^{(t)})\right]+\frac{L}{2}\mathbb{E}\left[ \|U_i^{(t)} - W_i^{(t)}\|_F^2\right]+\eta_{i,t}^2C_AC_BG^3 \nonumber \\
        &\qquad- \eta_{i,t}(\mathbb{E}\left[\|\nabla_B L_i(W_i^{{(t)}})\|_F^2\right]+\mathbb{E}\left[\|\nabla_A L_i(W_i^{{(t)}})\|_F^2\right])
        \label{L_smooth_UW2}
    \end{align}
    If $t+1\in \mathcal{I}_E$, by lemma~\ref{lemma_innerproduct} and Eq.~(\ref{L_smooth_UW}) and~(\ref{L_smooth_VU}), we have:
    \begin{align}
        \mathbb{E}\left[\mathcal{L}_i(W_i^{(t+1)})\right] \nonumber
        &= \mathbb{E}\left[\mathcal{L}_i(V_i^{(t)})\right]\nonumber\\
        &\le \mathbb{E}\left[\mathcal{L}_i(U_i^{(t)})\right]+\mathbb{E}\left[\langle V_i^{(t)} - U_i^{(t)},\nabla_W\mathcal{L}_i(U_i^{(t)}) \rangle_F\right] + \frac{L}{2}\mathbb{E}\left[\|V_i^{(t)} - U_i^{(t)}\|_F^2\right]\nonumber\\
        &\le \mathbb{E}\left[\mathcal{L}_i(W_i^{(t)})\right]+\mathbb{E}\left[\langle U_i^{(t)}-W_i^{(t)},\nabla_W\mathcal{L}_i(W_i^{(t)}) \rangle_F\right]+\frac{L}{2}\mathbb{E}\left[ \|U_i^{(t)} - W_i^{(t)}\|_F^2\right]\nonumber\\
        & \qquad +\mathbb{E}\left[\langle V_i^{(t)} - U_i^{(t)},\nabla_W\mathcal{L}_i(U_i^{(t)}) \rangle_F\right] + \frac{L}{2}\mathbb{E}\left[\|V_i^{(t)} - U_i^{(t)}\|_F^2\right] \nonumber\\
        &\le \mathbb{E}\left[\mathcal{L}_i(W_i^{(t)})\right]+\frac{L}{2}\mathbb{E}\left[ \|U_i^{(t)} - W_i^{(t)}\|_F^2\right]+\eta_{i,t}^2C_AC_BG^3 \nonumber\\
        &\qquad - \eta_{i,t}(\mathbb{E}\left[\|\nabla_B L_i(W_i^{{(t)}})\|_F^2\right]+\mathbb{E}\left[\|\nabla_A L_i(W_i^{{(t)}})\|_F^2\right])\nonumber\\
        & \qquad +\mathbb{E}\left[\left|\langle V_i^{(t)} - U_i^{(t)},\nabla_W\mathcal{L}_i(U_i^{(t)}) \rangle_F\right|\right] + \frac{L}{2}\mathbb{E}\left[\|V_i^{(t)} - U_i^{(t)}\|_F^2\right]
        \label{L_smooth_VW}
    \end{align}
    By comparing Eq.~(\ref{L_smooth_UW2}) and Eq.~(\ref{L_smooth_VW}), we observe that Equation (6) holds universally for arbitrary values of $t$. So we get, for $\forall t$ and $\eta_{i,j} = \eta$,
    \begin{align}
        \mathbb{E}\left[\mathcal{L}_i(W_i^{(t+1)})\right]
        &\le \mathbb{E}\left[\mathcal{L}_i(W_i^{(t)})\right]+\frac{L}{2}\mathbb{E}\left[ \|U_i^{(t)} - W_i^{(t)}\|_F^2\right]+\eta^2C_AC_BG^3 \nonumber\\
        &\qquad - \eta(\mathbb{E}\left[\|\nabla_B L_i(W_i^{{(t)}})\|_F^2\right]+\mathbb{E}\left[\|\nabla_A L_i(W_i^{{(t)}})\|_F^2\right])\nonumber\\
        & \qquad +\mathbb{E}\left[\left|\langle V_i^{(t)} - U_i^{(t)},\nabla_W\mathcal{L}_i(U_i^{(t)}) \rangle_F\right|\right] + \frac{L}{2}\mathbb{E}\left[\|V_i^{(t)} - U_i^{(t)}\|_F^2\right]
    \end{align}
    Moreover, we have:
    \begin{align}
        \eta(\mathbb{E}\left[\|\nabla_B L_i(W_i^{{(t)}})\|_F^2\right]+\mathbb{E}\left[\|\nabla_A L_i(W_i^{{(t)}})\|_F^2\right])
        &\le \mathbb{E}\left[\mathcal{L}_i(W_i^{(t)}) - \mathcal{L}_i(W_i^{t+1})\right]+\frac{L}{2}\mathbb{E}\left[ \|U_i^{(t)} - W_i^{(t)}\|_F^2\right]+\eta^2C_AC_BG^3\nonumber\\
        & \qquad +\mathbb{E}\left[\left|\langle V_i^{(t)} - U_i^{(t)},\nabla_W\mathcal{L}_i(U_i^{(t)}) \rangle_F\right|\right] + \frac{L}{2}\mathbb{E}\left[\|V_i^{(t)} - U_i^{(t)}\|_F^2\right]
        \label{gradient_aprox+local}
    \end{align}
    Since:
    \begin{align}
        \langle  V_i^{(t)} - U_i^{(t)},\nabla_W\mathcal{L}_i(U_i^{(t)}) \rangle_F 
        &\le \frac{1}{2}\left[\frac{1}{\eta}\|V_i^{(t)}-U_i^{(t)}\|_F^2+\eta\|\nabla_W\mathcal{L}_i(U_i^{(t)})\|_F^2  \right]\nonumber\\
        &\le\frac{1}{2\eta}\|V_i^{(t)} - U_i^{(t)}\|_F^2 + \frac{\eta}{2}G^2
        \label{innerproduct_VU}
    \end{align}
    The last inequality in~\ref{innerproduct_VU} holds by Assumption~\ref{Assumption_gradient-bounded}. By Eq.~(\ref{gradient_aprox+local}) and~(\ref{innerproduct_VU}), we have:
    \begin{align}
        \mathbb{E}\left[\|\nabla_B L_i(W_i^{{(t)}})\|_F^2\right]+\mathbb{E}\left[\|\nabla_A L_i(W_i^{{(t)}})\|_F^2\right]
        &\le \frac{\mathbb{E}\left[\mathcal{L}_i(W_i^{(t)}) - \mathcal{L}_i(W_i^{t+1})\right]}{\eta} + \frac{1}{\eta}(\frac{L}{2}\mathbb{E}\left[ \|U_i^{(t)} - W_i^{(t)}\|_F^2\right]\nonumber\\
        &\qquad +\eta^2C_AC_BG^3+(\frac{1}{2\eta}+\frac{L}{2})\mathbb{E}\left[\|V_i^{(t)}-U_i^{(t)}\|_F^2\right]+\frac{\eta}{2}G^2)
        \label{gradient_aprox_local2}
    \end{align}
    Summing Eq.~(\ref{gradient_aprox_local2}) over $i=1$ to $m$, by lemma.~\ref{lemma_dist_UW} and the convergence condition in Definition.~\ref{Weak_Convergence_Condition}, Eq.~(\ref{gradient_aprox_local2}) can be transformed to:
    \begin{align}
        &\frac{1}{m}\sum_{i=1}^m(\mathbb{E}\left[\|\nabla_B L_i(W_i^{{(t)}})\|_F^2\right]+\mathbb{E}\left[\|\nabla_A L_i(W_i^{{(t)}})\|_F^2\right])\nonumber\\ 
        &\le \frac{1}{m}\sum_{i=1}^m\frac{\mathbb{E}\left[\mathcal{L}_i(W_i^{(t)}) - \mathcal{L}_i(W_i^{(t+1)})\right]}{\eta} + \frac{1}{\eta}(\frac{L}{2}\cdot \frac{1}{m}\sum_{i=1}^m\mathbb{E}\left[ \|U_i^{(t)} - W_i^{(t)}\|_F^2\right]\nonumber\\
        &\qquad +\eta^2C_AC_BG^3+(\frac{1}{2\eta}+\frac{L}{2})\cdot\frac{1}{m}\sum_{i=1}^m\mathbb{E}\left[\|V_i^{(t)}-U_i^{(t)}\|_F^2\right]+\frac{\eta}{2}G^2)\nonumber\\
        &\le \frac{1}{m}\sum_{i=1}^m\frac{\mathbb{E}\left[\mathcal{L}_i(W_i^{(t)}) - \mathcal{L}_i(W_i^{(t+1)})\right]}{\eta} + \frac{1}{\eta}(\frac{L}{2}(3\eta^2C_A^2C_B^2G^4+3C_A^4G^2\nonumber\\
        &\qquad +3C_B^4G^2)\eta^2 + \eta^2C_AC_BG^3+(\frac{1}{2\eta}+\frac{L}{2})R^2\eta^2 + \frac{\eta}{2}G^2)\nonumber\\
        &\le \frac{1}{m}\sum_{i=1}^m\frac{\mathbb{E}\left[\mathcal{L}_i(W_i^{(t)}) - \mathcal{L}_i(W_i^{(t+1)})\right]}{\eta} + \frac{1}{\eta} \cdot M\eta^2\nonumber\\
        &\le \frac{1}{m}\sum_{i=1}^m\frac{\mathbb{E}\left[\mathcal{L}_i(W_i^{(t)}) - \mathcal{L}_i(W_i^{(t+1)})\right]}{\eta} + M\eta
        \label{gradient_aprox_local3} 
    \end{align}
    Where $\frac{3}{2}L(\eta^2C_A^2C_B^2G^4+C_A^4G^2+C_B^2G^2)\eta^2+C_AC_BG^3\eta^2+\frac{L}{2}R^2\eta^2+\frac{1}{2}(R^2+G^2)\eta\le M\eta^2$, it holds if there exist some constant $\epsilon$ such that $\eta>\epsilon>0$. The second inequality holds by lemma.~\ref{lemma_dist_UW} and the convergence condition in Definition.~\ref{Weak_Convergence_Condition}. Summing Eq.~(\ref{gradient_aprox_local3}) over $t=1$ to $T$ yields:
    \begin{align}
        \frac{1}{mT}\sum_{t=1}^T\sum_{i=1}^m(\mathbb{E}\left[\|\nabla_B L_i(W_i^{{(t)}})\|_F^2\right]+\mathbb{E}\left[\|\nabla_A L_i(W_i^{{(t)}})\|_F^2\right])
        &\le \frac{1}{m}\sum_{i=1}^m\frac{\mathbb{E}\left[\mathcal{L}_i(W_i^{(0)}) - \mathcal{L}_i(W_i^{(T+1)})\right]}{\eta T}+M\eta\nonumber\\
        &\le \frac{1}{m}\sum_{i=1}^m\frac{\mathbb{E}\left[\mathcal{L}_i(W_i^{(0)}) - \mathcal{L}_i(W_i^*)\right]}{\eta T} + M\eta\nonumber\\
        &\le \frac{D}{\eta T}+M\eta
        \label{gradient_aprox_local4}
    \end{align}
    Where $W_i^* = \text{argmin}_{W}\mathcal{L}_i(W)$, $W_i^{(0)} = W_0$ and we assume that $\mathcal{L}_i(W_0) - \mathcal{L}_i(W_i^*)\le D$ for $\forall i$ . Based on the preceding reasoning and let $\eta = \sqrt{\frac{D}{MT}}$, we arrive at the following inequality:
    \begin{align}
        \frac{1}{mT}\sum_{i=1}^m\sum_{t=1}^T(\mathbb{E}\left[\|\nabla_B L_i(W_i^{{(t)}})\|_F^2\right]+\mathbb{E}\left[\|\nabla_A L_i(W_i^{{(t)}})\|_F^2\right])
        \le 2\sqrt{\frac{DM}{T}}
        \label{gradient_aprox_local5}
    \end{align}
\end{proof}

\subsection{Proof of lemmas}
Though Lemmas~\ref{lemma_innerproduct} and~\ref{lemma_dist_UW} replicate the proof methodology of \cite{guo2025selective}, we retain their proofs in this subsection to ensure a self-contained theoretical presentation, particularly given their critical role in deriving Theorem~\ref{thm_aggregation}.
\subsubsection{Proof of lemma~\ref{lemma_innerproduct}}
\begin{proof}
    We know that:
    \begin{align}
        U_i^{(t)} - W_i^{(t)}
        &= B_i^{(t+1)} A_i^{(t+1)} - B_i^{(t)} A_i^{(t)} \nonumber\\
        &=\left( B_i^{(t)} - \eta_{i,t} \nabla_B \mathcal{L}_i(W_i^{(t)}, \xi_{i,t}) \right)\left( A_i^{(t)} - \eta_{i,t} \nabla_A \mathcal{L}_i(W_i^{(t)}, \xi_{i,t}) \right) - B_i^{(t)} A_i^{(t)}\nonumber\\
        &= \eta_{i,t}^2 \nabla_W \mathcal{L}_i(W_i^{(t)}, \xi_{i,t}) A_i^{(t)\top} B_i^{(t)\top}
        \nabla_W \mathcal{L}_i(W_i^{(t)}, \xi_{i,t})  - \eta_{i,t} \nabla_B \mathcal{L}_i(W_i^{(t)},\xi_{i,t})A_i^{(t)} \nonumber\\
        &\qquad - \eta_{i,t} B_i^{(t)}\nabla_A \mathcal{L}_i(W_i^{(t)}, \xi_{i,t})
        \label{difference_UW}
\end{align}
Where the third equation is from $\nabla_B \mathcal{L}_i(W_i^{(t)}, \xi_{i,t}) = \nabla_W \mathcal{L}_i(W_i^{(t)}, \xi_{i,t}) A_i^{(t)\top}$ and $\nabla_A \mathcal{L}_i(W_i^{(t)}, \xi_{i,t})=B_i^{(t)\top} \nabla_W \mathcal{L}_i(W_i^{(t)}, \xi_{i,t})$ since $W_i^{(t)} = W_0+B_i^{(t)}A_i^{(t)}$. Then we have:
\begin{align}
    \mathbb{E} \left\langle U_i^{(t)} - W_i^{(t)}, \nabla_W \mathcal{L}_i(W_i^{(t)}) \right\rangle_F
    &= \eta_{i,t}^2 \mathbb{E} \left\langle\nabla_W \mathcal{L}_i(W_i^{(t)}, \xi_{i,t}) A_i^{(t)\top}B_i^{(t)\top}\nabla_W \mathcal{L}_i(W_i^{(t)}, \xi_{i,t}), \nabla_W \mathcal{L}_i(W_i^{(t)})\right\rangle_F\nonumber\\
    &\quad - \eta_{i,t} \mathbb{E} \left\langle \nabla_B \mathcal{L}_i(W_i^{(t)}, \xi_{i,t}) A_i^{(t)}, \nabla_W \mathcal{L}_i(W_i^{(t)})\right\rangle_F \nonumber\\
    &\quad - \eta_{i,t} \mathbb{E} \left\langle B_i^{(t)}\nabla_A \mathcal{L}_i(W_i^{(t)}, \xi_{i,t}), \nabla_W \mathcal{L}_i(W_i^{(t)})\right\rangle_F \nonumber\\
    &= \eta_{i,t}^2 \left\langle \nabla_W \mathcal{L}_i(W_i^{(t)}) A_i^{(t)\top}B_i^{(t)\top} \nabla_W \mathcal{L}_i(W_i^{(t)}), \nabla_W \mathcal{L}_i(W_i^{(t)})\right\rangle_F\nonumber \\
    &\quad - \eta_{i,t} \left\langle \nabla_B \mathcal{L}_i(W_i^{(t)}), \nabla_W\mathcal{L}_i(W_i^{(t)})A_i^{(t)\top}\right\rangle_F \nonumber\\
    &\quad - \eta_{i,t} \left\langle \nabla_A \mathcal{L}_i(W_i^{(t)}), B_i^{(t)\top}\nabla_W\mathcal{L}_i(W_i^{(t)})\right\rangle_F
    \label{proof_lemma1_innerproduct}
\end{align}
By Assumption~\ref{Assumption_gradient-bounded} and~\ref{Assumption_norm_bounded}, we have:
\begin{align}
    \left\langle \nabla_W \mathcal{L}_i(W_i^{(t)}) A_i^{(t)\top}B_i^{(t)\top} \nabla_W \mathcal{L}_i(W_i^{(t)}), \nabla_W \mathcal{L}_i(W_i^{(t)})\right\rangle_F 
    &\le \|\nabla_W \mathcal{L}_i(W_i^{(t)}) A_i^{(t)\top}B_i^{(t)\top} \nabla_W \mathcal{L}_i(W_i^{(t)})\|_F\cdot\|\nabla_W \mathcal{L}_i(W_i^{(t)})\|_F\nonumber\\
    &\le C_AC_BG^3 
    \label{proof_lemma1_part1}
\end{align}
By $\nabla_B \mathcal{L}_i(W_i^{(t)}, \xi_{i,t}) = \nabla_W \mathcal{L}_i(W_i^{(t)}, \xi_{i,t}) A_i^{(t)\top}$ and $\nabla_A \mathcal{L}_i(W_i^{(t)}, \xi_{i,t})=B_i^{(t)\top} \nabla_W \mathcal{L}_i(W_i^{(t)}, \xi_{i,t})$, we have:
\begin{align}
     \left\langle \nabla_B \mathcal{L}_i(W_i^{(t)}), \nabla_W\mathcal{L}_i(W_i^{(t)})A_i^{(t)\top}\right\rangle_F = \|\nabla_B \mathcal{L}_i(W_i^{(t)})\|_F^2
    \label{proof_lemma1_part2}
\end{align}
and
\begin{align}
    \left\langle \nabla_A \mathcal{L}_i(W_i^{(t)}), B_i^{(t)\top}\nabla_W\mathcal{L}_i(W_i^{(t)})\right\rangle_F=\|\nabla_A \mathcal{L}_i(W_i^{(t)})\|_F^2
    \label{proof_lemma1_part3}
\end{align}
Finally, we can get the result by Equation~(\ref{proof_lemma1_innerproduct}),~(\ref{proof_lemma1_part1}),~(\ref{proof_lemma1_part2}),~(\ref{proof_lemma1_part3}):
\begin{align}
    \mathbb{E}\left[\langle U_i^{(t)}-W_i^{(t)}, \nabla_W\mathcal{L}_i(W_i^{(t)}) \rangle_F\right] \le \eta_{i,t}^2C_AC_BG^3 - \eta_{i,t}(\mathbb{E}\left[\|\nabla_B L_i(W_i^{{(t)}})\|_F^2\right]+\mathbb{E}\left[\|\nabla_A L_i(W_i^{{(t)}})\|_F^2\right])
\end{align}
\end{proof}

\subsubsection{Proof of lemma~\ref{lemma_dist_UW}}
\begin{proof}
    By Equation~(\ref{difference_UW}), Assumption~\ref{Assumption_gradient-bounded} and~\ref{Assumption_norm_bounded}, we can get:
    \begin{align}
        \mathbb{E}\left[\|U_i^{(t)} - W_i^{(t)}\|_F^2\right]
        &=\mathbb{E}\| \eta_{i,t}^2 \nabla_W \mathcal{L}_i(W_i^{(t)}, \xi_{i,t}) A_i^{(t)\top} B_i^{(t)\top}
        \nabla_W \mathcal{L}_i(W_i^{(t)}, \xi_{i,t}) A_i^{(t)\top} - \eta_{i,t} \nabla_W \mathcal{L}_i(W_i^{(t)},\xi_{i,t}) A_i^{(t)\top}A_i^{(t)} \nonumber\\
        &\qquad - \eta_{i,t} B_i^{(t)} B_i^{(t)\top} \nabla_W \mathcal{L}_i(W_i^{(t)}, \xi_{i,t})\|_F^2\nonumber\\
        &\le 3\mathbb{E}\left[\|\eta_{i,t}^2 \nabla_W \mathcal{L}_i(W_i^{(t)}, \xi_{i,t}) A_i^{(t)\top} B_i^{(t)\top}
        \nabla_W \mathcal{L}_i(W_i^{(t)}, \xi_{i,t}) \|_F^2\right] \nonumber\\
        &\qquad +3\mathbb{E}\left[\|\eta_{i,t} \nabla_W \mathcal{L}_i(W_i^{(t)},\xi_{i,t}) A_i^{(t)\top}A_i^{(t)}\|_F^2\right]+3\mathbb{E}\left[\|\eta_{i,t} B_i^{(t)} B_i^{(t)\top} \nabla_W \mathcal{L}_i(W_i^{(t)}, \xi_{i,t})\|_F^2\right]\nonumber\\
        &\le 3\eta_{i,t}^4C_A^2C_B^2G^4 + 3\eta_{i,t}^2C_A^4G^2+3\eta_{i,t}^2C_B^4G^2
    \end{align}
\end{proof}

\subsection{Proof of Corollary~\ref{corollary_weak_optimal}}\label{Appendix_pf_corollary1}
\begin{proof}
    First, we show that if the Aggregation-Broadcast Operators(ABO) $\mathcal{P}$ and $\mathcal{Q}$, can satisfy the convergence condition in Definition~\ref{Weak_Convergence_Condition} if:
    \begin{align}
        \mathcal{P}(A_{1\le j\le m}^{(t+1)},B_{1\le j\le m}^{(t+1)})\mathcal{Q}(A_{1\le j\le m}^{(t+1)},B_{1\le j\le m}^{(t+1)}) = \frac{1}{m}\sum_{j=1}^m B_j^{(t+1)}A_j^{(t+1)}
        \label{pf_of_col1_1}
    \end{align}
    It is know by Eq.~(\ref{pf_of_col1_1}) that:
    \begin{align}
        &\frac{1}{m}\sum_{i=1}^m\|\mathcal{P}(A_{1\le j\le m}^{(t+1)},B_{1\le j\le m}^{(t+1)})\mathcal{Q}(A_{1\le j\le m}^{(t+1)},B_{1\le j\le m}^{(t+1)})-B_i^{(t+1)}A_i^{(t+1)}\|_F^2 \nonumber\\
        &= \frac{1}{m}\sum_{i=1}^m\|\frac{1}{m}\sum_{j=1}^m (B_j^{(t+1)}A_j^{(t+1)}-B_i^{(t+1)}A_i^{(t+1)})\|_F^2\nonumber\\
        &\le \frac{1}{m^2}\sum_{i=1}^m\sum_{j=1}^m\|(B_j^{(t+1)}A_j^{(t+1)}-B_i^{(t+1)}A_i^{(t+1)})\|_F^2\nonumber\\
        &=\frac{1}{m^2}\sum_{i=1}^m\sum_{j=1}^m\|B_j^{(t+1)}(A_j^{(t+1)}-A_i^{(t+1)}) + (B_j^{(t+1)}-B_i^{(t+1)})A_i^{(t+1)}\|_F^2\nonumber\\
        &\le \frac{2}{m^2}\sum_{i=1}^m\sum_{j=1}^m(\|B_j^{(t+1)}(A_j^{(t+1)}-A_i^{(t+1)})\|_F^2+\|B_j^{(t+1)}-B_i^{(t+1)})A_i^{(t+1)}\|_F^2)\nonumber\\
        &\le \frac{2}{m^2}\sum_{i=1}^m\sum_{j=1}^m(C_B^2\cdot \|A_j^{(t+1)}-A_i^{(t+1)}\|_F^2+C_A^2\cdot \|B_j^{(t+1)}-B_i^{(t+1)}\|_F^2)
        \label{pf_of_col1_pq_ba}
    \end{align}
    Next, we consider about $\|A_j^{(t+1)}-A_i^{(t+1)}\|_F^2$ and $\|B_j^{(t+1)}-B_i^{(t+1)}\|_F^2$. If $t+1\in \mathcal{I}_E$, then there comes to a communication round, therefore $A_j^{(t+1)}=A_i^{(t+1)}$ and $B_j^{(t+1)}=B_i^{(t+1)}$ for $1\le i\le m$. On the other hand, if $t+1\notin \mathcal{I}_E$, we suppose that $nE<t+1<(n+1)E$ for an non-negative integer $n$. Then we have:
    \begin{align}
        A_j^{(t+1)} 
        &=A_j^{(nE)} -\sum_{t_0=nE}^t\eta\nabla_AL_j(W_j^{(t_0)},\xi_{j,t_0})\nonumber\\
        &=A_j^{(nE)} -\sum_{t_0=nE}^t\eta B_j^{(t_0)\top}\nabla_WL_j(W_j^{(t_0)},\xi_{j,t_0})
        \label{pf_of_col1_A_update}
    \end{align}
    Where $\xi_{i,t}$ is the i-th client's local data uniformly at random at the training step $t$. The second equation is from  $\nabla_A \mathcal{L}_i(W_i^{(t)}, \xi_{i,t})=B_i^{(t)\top} \nabla_W \mathcal{L}_i(W_i^{(t)}, \xi_{i,t})$. Then by Eq.~(\ref{pf_of_col1_A_update}) and Assumption~\ref{Assumption_norm_bounded} we have:
    \begin{align}
        \|A_j^{(t+1)}-A_i^{(t+1)}\|_F^2 
        &= \|(A_j^{(nE)} -\sum_{t_0=nE}^t\eta B_j^{(t_0)\top}\nabla_WL_j(W_j^{(t_0)},\xi_{j,t_0})) \nonumber\\
        &\qquad - (A_i^{(nE)} -\sum_{t_0=nE}^t\eta B_i^{(t_0)\top}\nabla_WL_i(W_i^{(t_0)},\xi_{i,t_0}))\|_F^2\nonumber\\
        &=\|\eta\sum_{t_0=nE}^t(B_j^{(t_0)\top}\nabla_WL_j(W_j^{(t_0)},\xi_{j,t_0})-B_i^{(t_0)\top}\nabla_WL_i(W_i^{(t_0)},\xi_{i,t_0}))\|_F^2\nonumber\\
        &\le \eta^2(t-nE+1)\sum_{t_0=nE}^t\|B_j^{(t_0)\top}\nabla_WL_j(W_j^{(t_0)},\xi_{j,t_0})-B_i^{(t_0)\top}\nabla_WL_i(W_i^{(t_0)},\xi_{i,t_0})\|_F^2\nonumber\\
        &\le \eta^2(t-nE+1)\sum_{t_0=nE}^t(2\|B_j^{(t_0)\top}\nabla_WL_j(W_j^{(t_0)},\xi_{j,t_0})\|_F^2+2\|B_i^{(t_0)\top}\nabla_WL_i(W_i^{(t_0)},\xi_{i,t_0})\|_F^2)\nonumber\\
        &\le \eta^2(t-nE+1)\sum_{t_0=nE}^t(2C_B^2\|L_j(W_j^{(t_0)},\xi_{j,t_0})\|_F^2+2C_B^2\|\nabla_WL_i(W_i^{(t_0)},\xi_{i,t_0})\|_F^2)
        \label{pf_of_col_diffA}
    \end{align}
    Take the expectation on Eq.~(\ref{pf_of_col_diffA}) and by Assumption~\ref{Assumption_gradient-bounded} we can get:
    \begin{align}
        \mathbb{E}\left[\|A_j^{(t+1)}-A_i^{(t+1)}\|_F^2\right] 
        &\le \eta^2(t-nE+1)\sum_{t_0=nE}^t(2C_B^2G^2+2C_B^2G^2)\nonumber\\
        &=4\eta^2(t-nE+1)^2C_B^2G^2\nonumber\\
        &\le 4E^2C_B^2G^2\eta^2
        \label{pf_of_col1_diffA2}
    \end{align}
    By the same reason, we have:
    \begin{align}
        \mathbb{E}\left[\|B_j^{(t+1)}-B_i^{(t+1)}\|_F^2\right]\le 4E^2C_A^2G^2\eta^2
        \label{pf_of_col1_diffB2}
    \end{align}
    Plugging Eq.~(\ref{pf_of_col1_diffA2}) and Eq.~(\ref{pf_of_col1_diffB2}) into Eq.~(\ref{pf_of_col1_pq_ba}) we can obtain:
    \begin{align}
        &\mathbb{E}\left[\frac{1}{m}\sum_{i=1}^m\|\mathcal{P}(A_{1\le j\le m}^{(t+1)},B_{1\le j\le m}^{(t+1)})\mathcal{Q}(A_{1\le j\le m}^{(t+1)},B_{1\le j\le m}^{(t+1)})-B_i^{(t+1)}A_i^{(t+1)}\|_F^2 \right]\nonumber\\
        &\le \frac{2}{m^2}\sum_{i=1}^m\sum_{j=1}^m(C_B^2\cdot 4E^2C_B^2G^2\eta^2+ C_A^2\cdot 4E^2C_A^2G^2\eta^2)\nonumber\\
        &=8E^2G^2(C_A^4+C_B^4)\eta^2
    \end{align}
    Let $R^2 = 8E^2G^2(C_A^4+C_B^4)$ we can proof that $\mathcal{P}$ and $\mathcal{Q}$ satisfy the convergence Condition in Definition~\ref{Weak_Convergence_Condition} if Eq~\ref{pf_of_col1_1} holds. Next, we proof that $\mathcal{P}$ and $\mathcal{Q}$ can achieve the optimal convergence rate if Eq~\ref{pf_of_col1_1} holds. Let $f(X) = \frac{1}{m}\sum_{i=1}^m\|X-X_i\|_F^2$, then we have:
    \begin{align}
        f(X) 
        &= \frac{1}{m}\sum_{i=1}^m\text{tr}((X-X_i)^\top(X-X_i))\nonumber\\
        &=\text{tr}(X^\top X) - 2\cdot \frac{1}{m}\sum_{i=1}^m\text{tr}(X_i^{\top}X) +\frac{1}{m}\text{tr}(X_i^{\top}X_i)
        \label{pf_of_col1_f}
    \end{align}
    Therefore:
    \begin{align}
        \nabla f(X) = 2X-\frac{2}{m}\sum_{i=1}^mX_i
    \end{align}
    It means that:
    \begin{align}
        \text{argmin}_Xf(X) = \text{argmin}_{X} \sum_{i=1}^m\|X-X_i\|_F^2 = \frac{1}{m}\sum_{i=1}^mX_i
        \label{pf_of_col1_argmin}
    \end{align}
    Finally, let $X = \mathcal{P}(A_{1\le j\le m}^{(t+1)},B_{1\le j\le m}^{(t+1)})\mathcal{Q}(A_{1\le j\le m}^{(t+1)},B_{1\le j\le m}^{(t+1)})$ and $X_i = B_i^{(t+1)}A_i^{(t+1)}$ we can get the proof.
\end{proof}
\subsection{Proof of Theorem~\ref{Strong_Convergence_thm}}\label{Appendix_pf_sc2}
\begin{proof}
    First, we show that the ABO $\mathcal{P}(A_{1\le j\le m}^{(t+1)},B_{1\le j\le m}^{(t+1)})$ and $\mathcal{Q}(A_{1\le j\le m}^{(t+1)},B_{1\le j\le m}^{(t+1)})$ satisty the weak convergence condition defiend in \ref{Weak_Convergence_Condition}.
    \begin{align}
         &\quad\frac{1}{m}\sum_{i=1}^m\|\mathcal{P}(A_{1\le j\le m}^{(t+1)},B_{1\le j\le m}^{(t+1)})\mathcal{Q}(A_{1\le j\le m}^{(t+1)},B_{1\le j\le m}^{(t+1)})-B_i^{(t+1)}A_i^{(t+1)}\|_F^2\nonumber\\
         &\le \frac{1}{m}\sum_{i=1}^m\|\mathcal{P}(A_{1\le j\le m}^{(t+1)},B_{1\le j\le m}^{(t+1)})(\mathcal{Q}(A_{1\le j\le m}^{(t+1)},B_{1\le j\le m}^{(t+1)})-A_i^{(t+1)})+(\mathcal{P}(A_{1\le j\le m}^{(t+1)},B_{1\le j\le m}^{(t+1)})-B_i^{(t+1)})A_i^{(t+1)}\|_F^2\nonumber\\
         &\le2\cdot\frac{1}{m}\sum_{i=1}^m\|\mathcal{P}(A_{1\le j\le m}^{(t+1)},B_{1\le j\le m}^{(t+1)})\|_F^2\cdot\|\mathcal{Q}(A_{1\le j\le m}^{(t+1)},B_{1\le j\le m}^{(t+1)})-A_i^{(t+1)}\|_F^2\nonumber\\
         &\quad+2\cdot\frac{1}{m}\sum_{i=1}^m\|\mathcal{P}(A_{1\le j\le m}^{(t+1)},B_{1\le j\le m}^{(t+1)})-B_i^{(t+1)})\|_F^2\cdot\|A_i^{(t+1)}\|_F^2
         \label{pf_of_thm3_1}
    \end{align}
    By the Eq.~(\ref{sum_PB}) and Assumption~\ref{Assumption_norm_bounded}, we can obtain the estimate of $\|\mathcal{P}(A_{1\le j\le m}^{(t+1)},B_{1\le j\le m}^{(t+1)})\|_F^2$:
    \begin{align}
        \|\mathcal{P}(A_{1\le j\le m}^{(t+1)},B_{1\le j\le m}^{(t+1)})\|_F^2 
        &=\|\frac{1}{m}\sum_{i=1}^m(\mathcal{P}(A_{1\le j\le m}^{(t+1)},B_{1\le j\le m}^{(t+1)})-B_i^{(t+1)})+\frac{1}{m}\sum_{i=1}^mB_i^{(t+1)}\|_F^2\nonumber\\
        &\le 2\|\frac{1}{m}\sum_{i=1}^m(\mathcal{P}(A_{1\le j\le m}^{(t+1)},B_{1\le j\le m}^{(t+1)})-B_i^{(t+1)})\|_F^2 + 2\|\frac{1}{m}\sum_{i=1}^mB_i^{(t+1)}\|_F^2\nonumber\\
        &\le \frac{2}{m}\sum_{i=1}^m\|\mathcal{P}(A_{1\le j\le m}^{(t+1)},B_{1\le j\le m}^{(t+1)})-B_i^{(t+1)}\|_F^2+\frac{2}{m}\sum_{i=1}^m\|B_i^{(t+1)}\|_F^2\nonumber\\
        &\le 2P^2\eta^2+2C_B^2
        \label{pf_of_thm3_2}
    \end{align}
    By combining Eq.~(\ref{pf_of_thm3_1}), Eq.~(\ref{pf_of_thm3_2}) and Assumption.~\ref{Assumption_norm_bounded} we can get:
    \begin{align}
        &\quad\frac{1}{m}\sum_{i=1}^m\|\mathcal{P}(A_{1\le j\le m}^{(t+1)},B_{1\le j\le m}^{(t+1)})\mathcal{Q}(A_{1\le j\le m}^{(t+1)},B_{1\le j\le m}^{(t+1)})-B_i^{(t+1)}A_i^{(t+1)}\|_F^2\nonumber\\
        &\le 2\cdot(2P^2\eta^2+2C_B^2)\cdot\frac{1}{m}\sum_{i=1}^m\|\mathcal{Q}(A_{1\le j\le m}^{(t+1)},B_{1\le j\le m}^{(t+1)})-A_i^{(t+1)}\|_F^2+2C_A^2P^2\eta^2 \nonumber\\
        &\le 4(P^2\eta^2+C_B^2)\cdot Q^2\eta^2+2C_A^2P^2\eta^2\nonumber\\
        &=4P^2Q^2\eta^4+4C_B^2Q^2\eta^2 +2C_A^2P^2\eta^2
        \label{pf_of_thm3_3}
    \end{align}
    Similarly, by the symmetry of $\mathcal{P}$ and $\mathcal{Q}$, we obtain:
    \begin{align}
        &\quad\frac{1}{m}\sum_{i=1}^m\|\mathcal{P}(A_{1\le j\le m}^{(t+1)},B_{1\le j\le m}^{(t+1)})\mathcal{Q}(A_{1\le j\le m}^{(t+1)},B_{1\le j\le m}^{(t+1)})-B_i^{(t+1)}A_i^{(t+1)}\|_F^2\nonumber\\
        &\le4P^2Q^2\eta^4+2C_B^2Q^2\eta^2 +4C_A^2P^2\eta^2
        \label{pf_of_thm3_4}
    \end{align}
    Take the average of Eq.~(\ref{pf_of_thm3_3}) and Eq.~(\ref{pf_of_thm3_4}) then yields:
    \begin{align}
        &\quad\frac{1}{m}\sum_{i=1}^m\|\mathcal{P}(A_{1\le j\le m}^{(t+1)},B_{1\le j\le m}^{(t+1)})\mathcal{Q}(A_{1\le j\le m}^{(t+1)},B_{1\le j\le m}^{(t+1)})-B_i^{(t+1)}A_i^{(t+1)}\|_F^2\nonumber\\
        &\le4P^2Q^2\eta^4+3C_B^2Q^2\eta^2 +3C_A^2P^2\eta^2\nonumber\\
        &= R^2\eta^2
    \end{align}
    Where $R^2 = 4P^2Q^2\eta^2+3C_B^2Q^2 +3C_A^2P^2$. By Definition~\ref{Weak_Convergence_Condition} we know that $\mathcal{P}$ and $\mathcal{Q}$ satisfy the weak convergence condition, then by Eq.~(\ref{gradient_aprox_local4}) we have:
    \begin{align}
        \frac{1}{mT}\sum_{t=1}^T\sum_{i=1}^m(\mathbb{E}\left[\|\nabla_B L_i(W_i^{{(t)}})\|_F^2\right]+\mathbb{E}\left[\|\nabla_A L_i(W_i^{{(t)}})\|_F^2\right])\le \frac{D}{\eta T}+M\eta
        \label{local_convergence_thm3}
    \end{align}
    
     It is worth noting that Eq.~(\ref{local_convergence_thm3}) provides an estimate of the convergence rate of the local models. However, the convergence of local models does not necessarily imply the convergence of the global model. Therefore, we next turn our attention to analyzing the convergence of the global loss $\mathcal{L}(W^{(t)}) = \frac{1}{m}\sum_{i=1}^m\mathcal{L}_i(W^{(t)})$ with global model parameter $W^{(t)} = W_0+\mathcal{P}(A_{1\le j\le m}^{(t)},B_{1\le j\le m}^{(t)})\mathcal{Q}(A_{1\le j\le m}^{(t)},B_{1\le j\le m}^{(t)})$ and $\nabla_B \mathcal{L}_i(W_i^{(t)}) = \nabla_W \mathcal{L}_i(W_i^{(t)}) A_i^{(t)\top}$, $\nabla_A \mathcal{L}_i(W_i^{(t)})=B_i^{(t)\top} \nabla_W \mathcal{L}_i(W_i^{(t)})$. We know that:
     \begin{align}
         \|\nabla_B\mathcal{L}(W^{(t)})\|_F^2 
         &\le \|\nabla_B\mathcal{L}(W^{(t)}) - \frac{1}{m}\sum_{i=1}^m \nabla_B\mathcal{L}_i(W_i^{(t)}) + \frac{1}{m}\sum_{i=1}^m\nabla_B\mathcal{L}_i(W_i^{(t)})\|_F^2\nonumber\\
         &\le 2\|\frac{1}{m}\sum_{i=1}^m\nabla_B\mathcal{L}_i(W^{(t)})-\frac{1}{m}\sum_{i=1}^m \nabla_B\mathcal{L}_i(W_i^{(t)})\|_F^2 +2\|\frac{1}{m}\sum_{i=1}^m \nabla_B\mathcal{L}_i(W_i^{(t)})\|_F^2\nonumber\\
         &\le \frac{2}{m}\sum_{i=1}^m\|\nabla_B\mathcal{L}_i(W^{(t)})-\nabla_B\mathcal{L}_i(W_i^{(t)})\|_F^2 + \frac{2}{m}\sum_{i=1}^m\|\nabla_B\mathcal{L}_i(W_i^{(t)})\|_F^2
         \label{gradient_B_1}
     \end{align}
     moreover, we can get:
     \begin{align}
         &\qquad \frac{1}{m}\sum_{i=1}^m\|\nabla_B\mathcal{L}_i(W^{(t)})-\nabla_B\mathcal{L}_i(W_i^{(t)})\|_F^2 \nonumber\\
         &= \frac{1}{m}\sum_{i=1}^m\|\nabla_W\mathcal{L}_i(W^{(t)})\mathcal{Q}(A_{1\le j\le m}^{(t)},B_{1\le j\le m}^{(t)})^\top - \nabla_W\mathcal{L}_i(W_i^{(t)})A_i^{(t)\top}\|_F^2\nonumber\\
         &=\frac{1}{m}\sum_{i=1}^m\|\nabla_W\mathcal{L}_i(W^{(t)})(\mathcal{Q}(A_{1\le j\le m}^{(t)},B_{1\le j\le m}^{(t)})^\top-A_i^{(t)\top})+(\nabla_W\mathcal{L}_i(W^{(t)}-\nabla_W\mathcal{L}_i(W_i^{(t)}))A_i^{(t)\top}\|_F^2\nonumber\\
         &\le\frac{2}{m}\sum_{i=1}^m\|\nabla_W\mathcal{L}_i(W^{(t)})(\mathcal{Q}(A_{1\le j\le m}^{(t)},B_{1\le j\le m}^{(t)})^\top-A_i^{(t)\top})\|_F^2 + \frac{2}{m}\sum_{i=1}^m\|\nabla_W\mathcal{L}_i(W^{(t)}-\nabla_W\mathcal{L}_i(W_i^{(t)}))A_i^{(t)\top}\|_F^2
         \label{gradient_B_2}
     \end{align}
     by Assumption~\ref{Assumption_gradient-bounded} and the strong convergence condition, we can get:
     \begin{align}
         &\qquad\frac{1}{m}\sum_{i=1}^m\|\nabla_W\mathcal{L}_i(W^{(t)})(\mathcal{Q}(A_{1\le j\le m}^{(t)},B_{1\le j\le m}^{(t)})^\top-A_i^{(t)\top}\|_F^2\nonumber\\
         &\le \frac{1}{m}\sum_{i=1}^m\|\nabla_W\mathcal{L}_i(W^{(t)})\|_F^2\cdot\|\mathcal{Q}(A_{1\le j\le m}^{(t)},B_{1\le j\le m}^{(t)})^\top-A_i^{(t)\top}\|_F^2\nonumber\\
         &\le G^2\cdot \frac{1}{m}\sum_{i=1}^m\|\mathcal{Q}(A_{1\le j\le m}^{(t)},B_{1\le j\le m}^{(t)})^\top-A_i^{(t)\top}\|_F^2\nonumber\\
         &\le G^2Q^2\eta^2
         \label{gradient_B_3}
     \end{align}
     by Assumption~\ref{Assumption_L_smooth}, Assumption~\ref{Assumption_norm_bounded} and Eq.~(\ref{local_convergence_thm3}), we have:
     \begin{align}
         &\qquad \frac{1}{m}\sum_{i=1}^m\|\nabla_W\mathcal{L}_i(W^{(t)}-\nabla_W\mathcal{L}_i(W_i^{(t)}))A_i^{(t)\top}\|_F^2\nonumber\\
         &\le \frac{1}{m}\sum_{i=1}^m\|\nabla_W\mathcal{L}_i(W^{(t)}-\nabla_W\mathcal{L}_i(W_i^{(t)}))\|_F^2 \cdot\|A_i^{(t)\top}\|_F^2\nonumber\\
         &\le \frac{1}{m}\sum_{i=1}^m C_A^2L^2\|W^{(t)}-W_i^{(t)}\|_F^2\nonumber\\
         &\le C_A^2L^2\cdot\frac{1}{m}\sum_{i=1}^m\|\mathcal{P}(A_{1\le j\le m}^{(t)},B_{1\le j\le m}^{(t)})\mathcal{Q}(A_{1\le j\le m}^{(t)},B_{1\le j\le m}^{(t)})-B_i^{(t)}A_i^{(t)}\|_F^2\nonumber\\
         &\le C_A^2L^2R^2\eta^2
         \label{gradient_B_4}
     \end{align}
     where $R^2 = 4P^2Q^2\eta^2+3C_B^2Q^2 +3C_A^2P^2$. Then by Eq.~(\ref{gradient_B_1}), Eq.~(\ref{gradient_B_2}), Eq.~(\ref{gradient_B_3}) and Eq.~(\ref{gradient_B_4}) we can get:
     \begin{align}
         \|\nabla_B\mathcal{L}(W^{(t)})\|_F^2\le 4G^2Q^2\eta^2+4C_A^2L^2R^2\eta^2 +\frac{2}{m}\sum_{i=1}^m\|\nabla_B\mathcal{L}_i(W_i^{(t)})\|_F^2
         \label{global_gradient_B}
     \end{align}
     for the same reason, we have:
     \begin{align}
         \|\nabla_A\mathcal{L}(W^{(t)})\|_F^2\le 4G^2P^2\eta^2+4C_B^2L^2R^2\eta^2+\frac{2}{m}\sum_{i=1}^m\|\nabla_A\mathcal{L}_i(W_i^{(t)})\|_F^2
         \label{global_gradient_A}
     \end{align}
     summing Eq.~(\ref{global_gradient_B}) and Eq.~(\ref{global_gradient_A}) over $t=1$ to $T$, then we can get:
     \begin{align}
         \frac{1}{T}\sum_{t=1}^T(\|\nabla_B\mathcal{L}(W^{(t)})\|_F^2+\|\nabla_A\mathcal{L}(W^{(t)})\|_F^2) 
         &\le 4G^2(Q^2+P^2)\eta^2 +4(C_A^2+C_B^2)L^2R^2\eta^2 \nonumber\\
         &\quad+2\cdot\frac{1}{mT}\sum_{t=1}^T\sum_{i=1}^m(\|\nabla_B\mathcal{L}_i(W_i^{(t)})\|_F^2+\|\nabla_A\mathcal{L}_i(W_i^{(t)})\|_F^2)
     \end{align}
     take the expectation on both side and by Eq.~(\ref{local_convergence_thm3}), we can get:
     \begin{align}
         &\qquad\frac{1}{T}\sum_{t=1}^T(\mathbb{E}\left[\|\nabla_B\mathcal{L}(W^{(t)})\|_F^2\right]+\mathbb{E}\left[\|\nabla_A\mathcal{L}(W^{(t)})\|_F^2\right])\nonumber\\
         &\le 4G^2(Q^2+P^2)\eta^2 +4(C_A^2+C_B^2)L^2R^2\eta^2+2(\frac{D}{\eta T}+M\eta)\nonumber\\
         &\le 2(\frac{D}{\eta T}+(M+N)\eta)
     \end{align}
     where we assume that $4G^2(Q^2+P^2)\eta^2 +4(C_A^2+C_B^2)L^2R^2\eta^2\le 2N\eta$, $\frac{3}{2}L(\eta^2C_A^2C_B^2G^4+C_A^4G^2+C_B^2G^2)\eta^2+C_AC_BG^3\eta^2+\frac{L}{2}R^2\eta^2+\frac{1}{2}(R^2+G^2)\eta\le M\eta^2$ and $\mathcal{L}_i(W_0) - \mathcal{L}_i(W_i^*)\le D$.
     let $\eta = \sqrt{\frac{D}{(M+N)T}}$, we can get the global model convergence rate as:
     \begin{align}
         \qquad\frac{1}{T}\sum_{t=1}^T(\mathbb{E}\left[\|\nabla_B\mathcal{L}(W^{(t)})\|_F^2\right]+\mathbb{E}\left[\|\nabla_A\mathcal{L}(W^{(t)})\|_F^2\right])\le 4\sqrt{\frac{D(M+N)}{T}}
     \end{align}
\end{proof}

\begin{subsection}{Proof of Corollary~\ref{col_Popt_Qopt}}\label{Proof_of_col4}
    \begin{proof}
        The proof of this corollary is similar to the proof of corollary~\ref{corollary_weak_optimal} which can be seen in Appendix~\ref{Appendix_pf_corollary1}. First, we chow that the Aggregation-Broadcast Operators(ABO) $\mathcal{P}$ and $\mathcal{Q}$ can satisfy the Sufficient  Condition in Theorem~\ref{sufficient_Condition2} if
        \begin{align}
         \mathcal{P}(A_{1\le j\le m}^{(t+1)},B_{1\le j\le m}^{(t+1)}) &= \frac{1}{m}\sum_{i=1}^m B_i^{(t+1)}
         \label{optmal_PB1}\\
         \mathcal{Q}(A_{1\le j\le m}^{(t+1)},B_{1\le j\le m}^{(t+1)}) &= \frac{1}{m}\sum_{i=1}^mA_i^{(t+1)}
         \label{optmal_QA1}
     \end{align}
     By Eq.~(\ref{optmal_PB1}) we know that:
     \begin{align}
         \mathbb{E}\left[\|\mathcal{P}(A_{1\le j\le m}^{(t+1)},B_{1\le j\le m}^{(t+1)})-B_i^{(t+1)}\|_F^2\right]
         &\le \mathbb{E}\left[\|\frac{1}{m}\sum_{j=1}^m(B_j^{(t+1)}-B_i^{(t+1)})\|_F^2\right]\nonumber\\
         &\le \frac{1}{m}\sum_{j=1}^m\mathbb{E}\left[\|B_j^{(t+1)} - B_i^{(t+1)}\|_F^2\right]\nonumber\\
         &\le 4E^2C_A^2G^2\eta^2
         \label{pf_of_col4_PB}
     \end{align}
     The last equation holds by Eq.~(\ref{pf_of_col1_diffB2}). Then by Eq.~(\ref{pf_of_col4_PB}) we can get:
     \begin{align}
         \mathbb{E}\left[\frac{1}{m}\sum_{i=1}^m\|\mathcal{P}(A_{1\le j\le m}^{(t+1)},B_{1\le j\le m}^{(t+1)})-B_i^{(t+1)}\|_F^2\right] 
         &\le \frac{1}{m}\sum_{i=1}^m\mathbb{E}\left[\|\mathcal{P}(A_{1\le j\le m}^{(t+1)},B_{1\le j\le m}^{(t+1)})-B_i^{(t+1)}\|_F^2\right]\nonumber\\
         & \le 4E^2C_A^2G^2\eta^2
     \end{align}
     Similarly, by Eq~\ref{pf_of_col1_diffA2} we have
     \begin{align}
         \mathbb{E}\left[\|\mathcal{Q}(A_{1\le j\le m}^{(t+1)},B_{1\le j\le m}^{(t+1)})-A_i^{(t+1)}\|_F^2\right]\le 4E^2C_B^2G^2\eta^2
         \label{pf_of_col4_QA}
     \end{align}
    and
    \begin{align}
        \mathbb{E}\left[\frac{1}{m}\sum_{i=1}^m\|\mathcal{Q}(A_{1\le j\le m}^{(t+1)},B_{1\le j\le m}^{(t+1)})-A_i^{(t+1)}\|_F^2\right] 
         &\le \frac{1}{m}\sum_{i=1}^m\mathbb{E}\left[\|\mathcal{Q}(A_{1\le j\le m}^{(t+1)},B_{1\le j\le m}^{(t+1)})-A_i^{(t+1)}\|_F^2\right]\nonumber\\
         & \le 4E^2C_B^2G^2\eta^2
    \end{align}
    Let $P^2 = 4E^2C_A^2G^2$ and $Q^2 = 4E^2C_B^2G^2$ we can proof that $\mathcal{P}$ and $\mathcal{Q}$ can satisfy the Sufficient  Condition in Theorem~\ref{sufficient_Condition2} under Eq.~(\ref{optmal_PB1}) and Eq.~(\ref{optmal_QA1}). Meanwhile, by the same step form Eq.~(\ref{pf_of_col1_f}) to Eq.~(\ref{pf_of_col1_argmin}), we can also proof that $\mathcal{P}$ and $\mathcal{Q}$ can achieve the optimal convergence rate of the global model.
    \end{proof}
\end{subsection}

\end{document}